%% file: main.tex
\documentclass[sigconf]{acmart} 

\AtBeginDocument{%
  }

\acmSubmissionID{6433}
\usepackage{colortbl} 
\definecolor{lightgray}{RGB}{230,230,230}
\definecolor{lightyellow}{RGB}{255,255,204}
\usepackage{bbm}  

\usepackage{microtype}
\usepackage{graphicx}
\usepackage{subfigure}
\usepackage{booktabs} 

\usepackage{hyperref}

\usepackage{algpseudocode}

\usepackage{amsmath}
\usepackage{mathtools}
\usepackage{amsthm}

\usepackage{url}         
\usepackage{algorithm}    
\usepackage{algpseudocode} 
\usepackage{subcaption}   
\usepackage{booktabs}     
\usepackage{array}        
\usepackage{tabularx}     
\usepackage[capitalize,noabbrev]{cleveref}

\theoremstyle{plain}
\newtheorem{theorem}{Theorem}[section]

\theoremstyle{definition}
\newtheorem{definition}[theorem]{Definition}
\newtheorem{assumption}[theorem]{Assumption}
\theoremstyle{remark}

\usepackage[textsize=tiny]{todonotes}

\usepackage{multirow}

\setcopyright{acmlicensed}
\copyrightyear{2025}
\acmYear{2025}
\setcopyright{cc}
\setcctype{by-nc-sa}
\acmConference[MM '25]{Proceedings of the 33rd ACM International Conference on Multimedia}{October 27--31, 2025}{Dublin, Ireland}
\acmBooktitle{Proceedings of the 33rd ACM International Conference on Multimedia (MM '25), October 27--31, 2025, Dublin, Ireland}\acmDOI{10.1145/3746027.3755856}
\acmISBN{979-8-4007-2035-2/2025/10}

\begin{document}

\title{Quantization Meets OOD: Generalizable Quantization-aware Training from a Flatness Perspective}

\author{Jiacheng Jiang}
\affiliation{%
  \institution{SIGS, Tsinghua University}
  \city{Shenzhen}
  \country{China}
}
\email{jiangjc23@mails.tsinghua.edu.cn}

\author{Yuan Meng$^\dagger$}
\affiliation{%
  \institution{Key Laboratory of Pervasive Computing, Ministry of Education \\
  Department of Computer Science and Technology, Tsinghua University}
  \city{Beijing}
  \country{China}
}
\email{yuanmeng@mail.tsinghua.edu.cn}

\author{Chen Tang}
\affiliation{%
  \institution{MMLab, The Chinese University of
 Hong Kong}
  \city{Hong Kong}
  \country{China}
}

\author{Han Yu}
\affiliation{%
  \institution{Department of Computer Science and Technology, Tsinghua University}
  \city{Beijing}
  \country{China}
}

\author{Qun Li}
\affiliation{%
  \institution{SIGS, Tsinghua University}
  \city{Shenzhen}
  \country{China}
}

\author{Zhi Wang$^\dagger$}
\affiliation{%
  \institution{SIGS, Tsinghua University}
  \city{Shenzhen}
  \country{China}
}
\email{wang_zhi@tsinghua.edu.cn}

\author{Wenwu Zhu$^\dagger$}
\affiliation{%
  \institution{Key Laboratory of Pervasive Computing, Ministry of Education \\
  Department of Computer Science and Technology, Tsinghua University}
  \city{Beijing}
  \country{China}
}
\email{wwzhu@tsinghua.edu.cn}

\thanks{$^\dagger$Corresponding author}
\renewcommand{\shortauthors}{Jiacheng Jiang et al.}

\input{./abstract.tex}




\maketitle


\input{./intro.tex}

\input{./preliminary}

\input{./obs.tex}

\input{./method}

\input{./experiment}
\input{./related_work.tex}

\input{./CONCLUSION_AND_FUTURE_WORK.tex}

\clearpage
\begin{acks}
  We sincerely thank the anonymous reviewers from ICLR, ICML, and ACM MM for their valuable feedback and suggestions. We also thank our lab mates for their help in improving the manuscript. This work is supported in part by National Key Research and Development Project of China (Grant No. 2023YFF0905502), National Natural Science Foundation of China (Grant No. 92467204, 62472249 and 62402264), and Shenzhen Science and Technology Program (Grant No. JCYJ20220818101014030 and KJZD20240903102300001). 
\end{acks}

\bibliographystyle{ACM-Reference-Format}
\balance
\bibliography{MMbib}

\input{./appendix.tex}
\end{document}

%% file: abstract.tex
\begin{abstract}
Current quantization-aware training (QAT) methods primarily focus on enhancing the performance of quantized models on in-distribution (I.D) data, while overlooking the potential performance degradation on out-of-distribution (OOD) data. 
In this paper, we first substantiate this problem through rigorous experiment, showing that \textit{QAT can lead to a significant OOD generalization performance degradation}.
Further, we find the contradiction between the perspective that flatness of loss landscape gives rise to superior OOD generalization and the phenomenon that QAT lead to a sharp loss landscape, can cause the above problem. 
Therefore, we propose a flatness-oriented QAT method, FQAT, to achieve generalizable QAT. 
Specifically, i) FQAT introduces a layer-wise freezing mechanism to mitigate the gradient conflict issue between dual optimization objectives (i.e., vanilla QAT and flatness). 
ii) FQAT proposes an disorder-guided adaptive freezing algorithm to dynamically determines which layers to freeze at each training step, effectively addressing the challenges caused by interference between layers.
A gradient disorder metric is designed to help the algorithm identify unstable layers during training. 
Extensive experiments on influential OOD benchmark demonstrate the superiority of our method
over state-of-the-art baselines under both I.D and OOD image classification tasks. {\small\textbf{Code:} \href{https://github.com/JachinJiang/Quantization-Meets-OOD-Generalizable-Quantization-aware-Training-from-a-Flatness-Perspective}{\texttt{https://github.com/JachinJiang/Quantization\\-Meets-OOD}}}

\end{abstract}

%% file: intro.tex
\section{Introduction} 

Quantization is one of the most effective technique for compressing deep neural networks (DNNs) to enable efficient inference and on-device execution while preserving high accuracy of computer vision (CV) tasks~\citep{hubara2021accurate, pmlr-v119-nagel20a,zhou2016dorefa, choi2018pact, esser2019learned}. 
By transforming pre-trained weights and activations from the high-bit (a.k.a., full 
precision) format, such as FP32, to more compact low-bit formats, such as INT8, quantization significantly reduces power consumption and speeds up inference, making it ideal for deploying neural networks on resource-limited edge devices~\citep{esser2019learned,jouppi2017datacenter,qiu2016going}.
Quantization-aware training (QAT) simulates quantization during training, allowing the model to adapt to quantization effects and achieve better performace than post-training quantzation~\citep{hubara2021accurate, pmlr-v119-nagel20a}.


Despite the notable success of
QAT methods, 
the existing works all focus on retaining the performance of full-precision models on in-distribution (I.D) data~\citep{zhou2016dorefa,tang2022mixed,tang2024retraining,esser2019learned, liu2021sharpness,yang2024gwq,shin2023nipq, lee2021network}, which may result in the performance degradation on out-of-distribution (OOD) data. However, OOD data is common in edge applications, such as foggy pedestrian detection in autonomous driving~\citep{huval2015empirical,levinson2011towards}.
Unfortunately, to the best of our knowledge, no prior research has systematically explored this potential risk. In this work, we take the first step by addressing a critical question:


\begin{center}
    \emph{Does quantization degrade the OOD generalization performance of a full-precision CV model?}
\end{center}

We perform systematic analysis on a representative class of QAT methods that with learned scale factors~\cite{esser2019learned,lee2021network,jung2019learning} and arrived at the following conclusion: 
\textbf{existing QAT methods lead to a decline in the model's OOD generalization performance}
Furthermore, we identify a key inconsistency with a prior study~\citep{javed2024qt}, which we trace back to data leakage in their experimental setup.

\begin{figure}[t]
    \centering
    \vspace{0cm}
    \includegraphics[width=\linewidth]{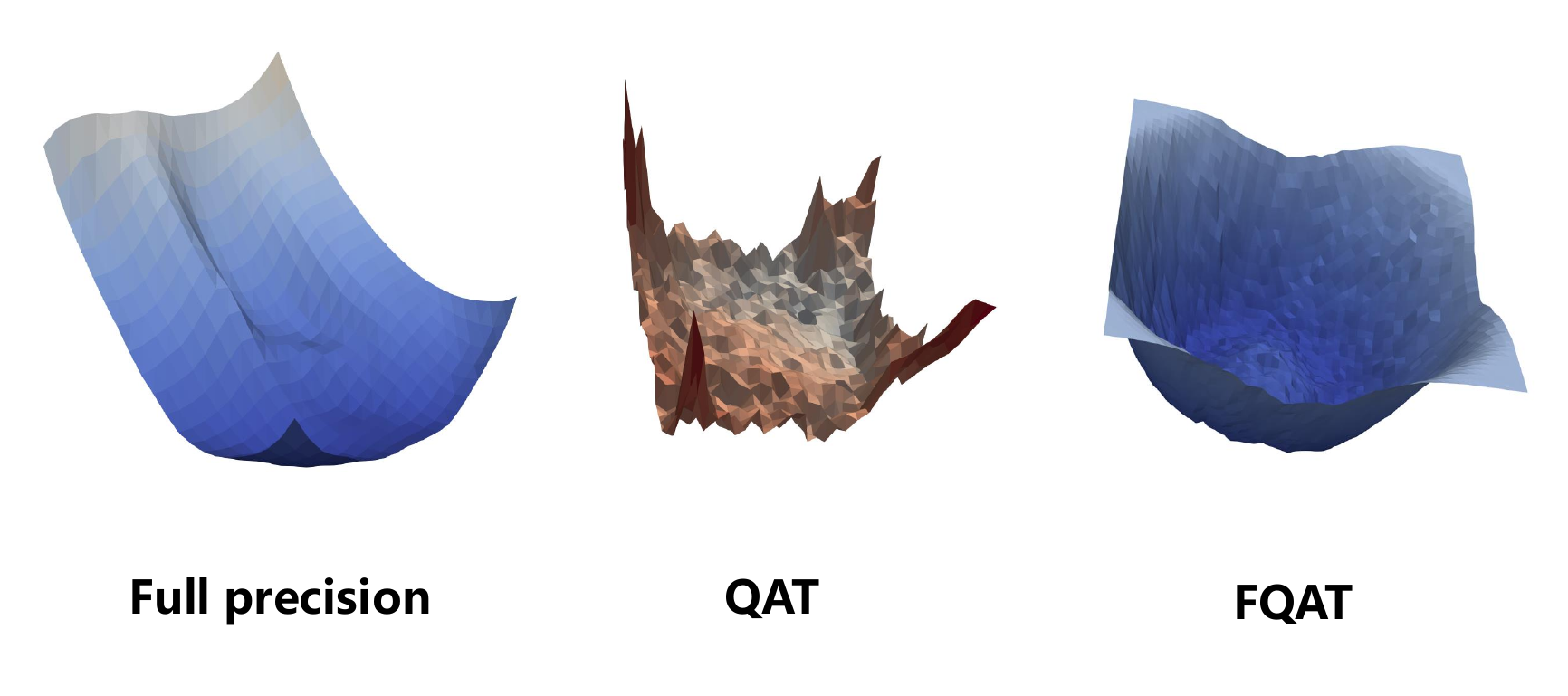}
    \caption{Visualization of loss landscapes of full precision model, quantized model (W4A4 with existing QAT method LSQ), quantized model (W4A4 with FQAT) on Sketch domain of PACS.}
    \label{fig:overloss} 
\end{figure}

Drawing inspiration from generalization techniques originally proposed for full-precision models—particularly sharpness-aware minimization (SAM)~\citep{foret2020sharpness}, we note that flatter loss landscapes are associated with improved OOD generalization~\citep{cha2021swad,wang2023sharpness,zhang2023flatness}.
This indicates, however, a sharp loss landscape induced by current QAT methods~\citep{liu2021sharpness}, which likely accounts for the observed performance degradation. 
Fig.~\ref{fig:overloss} visualizes the loss landscape of the full-precision model and the quantified model. The loss landscape of the quantified model is sharp compared to the full-precision model.

Based on the aforementioned findings, we propose a flatness-aware quantization-aware training method, FQAT, which incorporated a flatness-oriented optimization objective into the vanilla quantization-aware training objective to achieve generalizable quantized DNNs.
However, the optimization
of FQAT is non-trivial due to the following challenges.
First, it is challenging to optimize the key quantization parameter, \textit{scale factor}, due to its involvement in two optimization functions (i.e., vanilla QAT and flatness) with conflicting gradients.
We find that the flatness-oriented gradient exhibits greater instability.
Therefore, we propose a layer-wise QAT-gradient freezing method to mitigate the instability introduced by quantization.
Second, it is challenging to formulate the freezing policy.
The existing
freezing methods~\citep{nagel2022overcoming, tang2024retraining, liu2023oscillation} fail to consider the 
interference between scale factors of different layers under two optimization objectives, leading to poor generalization performances.
To tackle these challenges, we design a gradient disorder metric to identify more stable layers during training.
Based on this metric, we propose an adaptive freezing algorithm that leverages historical gradient information to automatically select layers for freezing quantization gradients in the future, thereby ensuring stable training for the flatness objective.
The last figure of Fig.~\ref{fig:overloss} also visualizes how our approach can effectively improve flatness of loss landscape.

In summary, we have made the following contributions: 
\begin{itemize}
    \item  To the best of our knowledge, this is the first work to propose and substantiate that quantization damages the OOD performance of full-precision CV models.
    \item We propose FQAT, a generalizable QAT method designed from the perspective of flatness. FQAT introduces a layer-wise freezing mechanism to mitigate gradient conflicts between standard QAT and flatness objectives. Additionally, we develop a disorder-guided adaptive freezing algorithm, which dynamically adjusts the layers to freeze at each training step based on historical gradient disorder.
    \item Extensive experiments on PACS, OfficeHome and DomainNet validate the effectiveness of FQAT. 
    Compared to the baselines, FQAT outperforms across all experimental settings and even surpasses the full-precision model in 8-bit quantization.
    FQAT improves both I.D 
    and OOD performance, with greater gains in OOD 
    (e.g., Under the 3-bit setting of PACS, OOD gain (+5.24\%) exceeds I.D gain (+2.61\%)).
    This highlights effectiveness of FQAT in enhancing OOD generalization.

\end{itemize}

%% file: preliminary.tex
\section{Preliminaries}
\label{gen_inst}



In this paper, we adopt the symmetric uniform quantization function for both weight and activation: 
\(
\hat{v} = Q(v; s) = s \times \left\lfloor \text{clip}\left( \frac{v}{s}, l, u \right) \right\rceil,
\)
where $v$ and $\hat{v}$ represent the full-precision value and the quantized value, respectively. 
The operator \(\lfloor \cdot \rceil\) denotes  rounding to the nearest integer, and \(s\) is a learnable scale factor in QAT~\citep{esser2019learned,tang2022mixed}.
The \(\texttt{clip}\) function ensures values stay within the bounds \([l, u]\). 
In \(b\)-bit quantization, for activation quantization, we set \(l = 0\) and \(u = 2^b - 1\); for weight quantization, we set \(l = -2^{b-1}\) and \(u = 2^{b-1} - 1\). 
Furthermore, to overcome the non-differentiability of the rounding operation, the Straight-Through Estimator (STE)~\citep{bengio2013estimating} is employed to approximate the gradients: \(
\frac{\partial \mathcal{L}}{\partial v} \approx \frac{\partial \mathcal{L}}{\partial \hat{v} } \cdot 1_{l \leq \frac{v}{s} \leq u},
\).

%% file: obs.tex
\section{OOD Generalization Assessment of Quantized DNNs }
This chapter will address the critical question posed in this paper and offer a detailed analysis.

\textbf{Experimental Settings.}
In assessment pipeline, the evaluated model is obtained through two steps: 1) obtaining a full-precision OOD generalizable model, 2)
performing quantization-oriented training.
Following the recent mainstream OOD model evaluation protocol~\citep{yu2024rethinking,zhang2023flatness, wang2023sharpness,cha2021swad}, we conduct step one by fine-tuning the self-supervised pretrained model with an OOD generalization method based on training dataset of the OOD benchmark.
The fine-tuning dataset is also used for QAT in step two.

There are three critical choices in the experiments: \textit{pre-trained model, full-precision OOD generalization method, and QAT method}. Previous OOD generalization research often employs ImageNet-pre-trained ResNet50 as the pre-trained model~\citep{ wang2023sharpness, zhang2023flatness,huang2020self}. However, a recent study~\citep{yu2024rethinking} highlights that ImageNet pre-training introduces severe data leakage issues, thereby compromising the validity of evaluations. Specifically, it becomes unclear whether performance improvements stem from retaining I.D performance or from genuine enhancements in OOD performance. To address this concern, MoCoV2-pre-trained ResNet50 is recommended as the pre-trained model~\citep{yu2024rethinking}. Furthermore, based on the sound pre-trained model, many OOD generalization methods underperform Empirical Risk Minimization (ERM)~\citep{vapnik1991principles,yu2024rethinking}. Therefore, we adopt ERM as our full-precision OOD generalization method. 
Finally, we select two representative QAT methods LSQ~\citep{esser2019learned} and EWGS~\citep{lee2021network} to handle the quantization process.
\begin{table}[t!]
\centering
\caption{LSQ results on three datasets. Experiments were run on seeds 0 and 23, with results reported as mean±std. LSQ: representative QAT method; LSQ+SAGM: Directly introducing SAGM objective to QAT; Ours: proposed FQAT; Val: I.D validation accuracy; Test: OOD test accuracy. The best performance is highlighted in bold.}
\vspace{-0.2cm}
\label{tab:combined_avg}
\scriptsize 
\renewcommand{\arraystretch}{1} 
\setlength{\tabcolsep}{3pt} 
\resizebox{0.48\textwidth}{!}{ 
\begin{tabular}{@{}llcccc@{}}
\toprule
\rowcolor{lightgray}
Dataset & Method & \multicolumn{1}{c}{Bit-width (W/A)} & Val & Test \\
\midrule

\multirow{12}{*}{DomainNet} 
  & ERM & Full & 62.04 & 40.95 \\
  \cmidrule(lr){2-5} 
  & LSQ & 4/4 & 60.95±0.38 & 39.14±0.24 \\
  & LSQ+SAGM & 4/4 & 61.79±0.26 & 40.07±0.23 \\
  & \cellcolor{lightyellow}Ours & \cellcolor{lightyellow}4/4 & \cellcolor{lightyellow}\textbf{62.52±0.25} & \cellcolor{lightyellow}\textbf{40.60±0.19} \\
  \cmidrule(lr){2-5}
  & LSQ & 5/5 & 60.43±0.37 & 38.46±0.16 \\
  & LSQ+SAGM & 5/5 & 62.73±0.42 & 40.51±0.37 \\
  & \cellcolor{lightyellow}Ours & \cellcolor{lightyellow}5/5 & \cellcolor{lightyellow}\textbf{63.19±0.36} & \cellcolor{lightyellow}\textbf{40.93±0.25} \\
  \cmidrule(lr){2-5}
  & LSQ & 8/8 & 60.97±0.31 & 39.10±0.16 \\
  & LSQ+SAGM & 8/8 & 63.21±0.36 & 41.10±0.35 \\
  & \cellcolor{lightyellow}Ours & \cellcolor{lightyellow}8/8 & \cellcolor{lightyellow}\textbf{63.39±0.21} & \cellcolor{lightyellow}\textbf{41.27±0.18} \\
\midrule
\multirow{12}{*}{PACS} 
  & ERM & Full & 96.42 & 85.29 \\
  \cmidrule(lr){2-5} 
  & LSQ & 3/3 & 77.39±0.68 & 54.35±2.23 \\
  & LSQ+SAGM & 3/3 & 74.98±1.49 & 50.60±1.37 \\
  & \cellcolor{lightyellow}Ours & \cellcolor{lightyellow}3/3 & 
  \cellcolor{lightyellow}\textbf{77.59±1.21} & \cellcolor{lightyellow}\textbf{55.84±1.48} \\
  \cmidrule(lr){2-5}
  & LSQ & 4/4 & 80.44±0.82 & 57.4±1.58 \\
  & LSQ+SAGM & 4/4 & 78.98±1.46 & 55.66±1.67 \\
  & \cellcolor{lightyellow}Ours & \cellcolor{lightyellow}4/4 & \cellcolor{lightyellow}\textbf{81.26±1.12} & \cellcolor{lightyellow}\textbf{59.42±1.25} \\
  \cmidrule(lr){2-5}
  & LSQ & 5/5 & 80.07±1.05 & 57.38±1.33 \\
  & LSQ+SAGM & 5/5 & 80.43±2.21 & 58.69±2.53 \\
  & \cellcolor{lightyellow}Ours & \cellcolor{lightyellow}5/5 & \cellcolor{lightyellow}\textbf{81.97±1.64} & \cellcolor{lightyellow}\textbf{60.41±2.43} \\
\midrule
\multirow{4}{*}{OfficeHome} 
  & ERM & Full & 78.44 & 60.31 \\
  \cmidrule(lr){2-5} 
  & LSQ & 3/3 & 59.58±2.07 & 38.15±1.85 \\
  & LSQ+SAGM & 3/3 & 60.69±1.75 & 39.98±1.72 \\
  & \cellcolor{lightyellow}Ours & \cellcolor{lightyellow}3/3 & \cellcolor{lightyellow}\textbf{62.12±1.11} & \cellcolor{lightyellow}\textbf{41.30±1.23} \\
\bottomrule
\end{tabular}
}
\vspace{-0.6cm}
\end{table}

\begin{table}[t!]
\centering
\caption{EWGS results on two datasets. Experiments were run on seeds 0 and 23, with results reported as mean±std. EWGS: representative QAT method; EWGS+SAGM: Directly introducing SAGM objective to QAT; Ours: proposed FQAT; Val: I.D validation accuracy; Test: OOD test accuracy. The best performance is highlighted in bold.}
\vspace{0.4cm}
\label{tab:combined_avg2}
\scriptsize 
\renewcommand{\arraystretch}{1} 
\setlength{\tabcolsep}{3pt} 
\resizebox{0.48\textwidth}{!}{ 
\begin{tabular}{@{}llcccc@{}}
\toprule
\rowcolor{lightgray}
Dataset & Method & \multicolumn{1}{c}{Bit-width (W/A)} & Val & Test \\
\midrule
\multirow{12}{*}{DomainNet} 
  & ERM & Full & 62.04 & 40.95 \\
  \cmidrule(lr){2-5} 
  & EWGS & 4/4 & 60.34±0.78 & 38.74±0.55 \\
  & EWGS+SAGM & 4/4 & 61.12±0.30 & 39.98±0.27 \\
  & \cellcolor{lightyellow}Ours & \cellcolor{lightyellow}4/4 & \cellcolor{lightyellow}\textbf{61.41±0.50} & \cellcolor{lightyellow}\textbf{40.23±0.34} \\
  \cmidrule(lr){2-5}
  & EWGS & 5/5 & 61.21±0.16 & 39.02±0.24 \\
  & EWGS+SAGM & 5/5 & 62.23±0.47 & 40.44±0.34 \\
  & \cellcolor{lightyellow}Ours & \cellcolor{lightyellow}5/5 & \cellcolor{lightyellow}\textbf{62.81±0.46} & \cellcolor{lightyellow}\textbf{40.88±0.26} \\
  \cmidrule(lr){2-5}
  & EWGS & 8/8 & 61.04±0.17 & 39.24±0.10 \\
  & EWGS+SAGM & 8/8 & 63.02±0.30 & 40.92±0.21 \\
  & \cellcolor{lightyellow}Ours & \cellcolor{lightyellow}8/8 & \cellcolor{lightyellow}\textbf{63.11±0.18} & \cellcolor{lightyellow}\textbf{41.01±0.16} \\
\midrule
\multirow{4}{*}{OfficeHome} 
  & ERM & Full & 78.44 & 60.31 \\
  \cmidrule(lr){2-5} 
  & EWGS & 3/3 & 60.32±2.41 & 39.09±2.93 \\
  & EWGS+SAGM & 3/3 & 60.12±1.21 & 39.12±1.51 \\
  & \cellcolor{lightyellow}Ours & \cellcolor{lightyellow}3/3 & \cellcolor{lightyellow}\textbf{62.13±1.40} & \cellcolor{lightyellow}\textbf{41.99±1.96} \\

\bottomrule
\end{tabular}
}
\vspace{-0.6cm}
\end{table}

\textbf{Conclusion: } \emph{Quantization can lead to OOD generalization performance degradation.}

In Table~\ref{tab:combined_avg} and ~\ref{tab:combined_avg2}, we present the experimental results on three OOD benchmarks: DomainNet, PACS and OfficeHome.
By comparing the experimental results of ERM and QAT methods across four different bit-widths, we find that quantization leads to significant performance
degradation.
It is worth mentioning that a related work~\citep{javed2024qt} has reached a conclusion inconsistent with ours.
We notice that they use ImageNet-pre-trained ResNet50~\citep{he2016deep} as pretrained model, which has been criticized for potential data leakage issues in OOD generalization evaluation~\cite{yu2024rethinking}.
Since the related work has not been open-sourced~\citep{javed2024qt}, we conducted ablation experiment within our own workflow by replacing the pre-trained model.
The 4-bit LSQ performance of Art domain in PACS improved from 51.07 (MoCoV2 pre-trained) to 76.51 (ImageNet pre-trained, without hyperparameter tuning), similar with the related work~\citep{javed2024qt}.
This result validates the soundness of our workflow and clarifies the inconsistency in conclusions.


SAM~\citep{cha2021swad,wang2023sharpness,zhang2023flatness} is an important research direction in the field of OOD generalization. It suggests that enhancing the flatness of the model's loss landscape can effectively improve its OOD generalization ability.
Specifically, flatness means that perturbations to the model's weights result in minimal fluctuations in its performance, thereby contributing to OOD generalization.
However, recent studies have revealed that quantization tends to sharpen the loss landscape~\citep{liu2021sharpness}. Based on these facts, we propose the following proposition and, in the next section, present an optimization method from the perspective of flatness.

\textbf{Proposition:} \emph{The sharp loss landscape caused by QAT may contribute to the degradation of OOD generalization ability.}

%% file: method.tex
\section{Method}
In this paper, we propose a flatness-oriented QAT (FQAT) method to maintain the OOD generalization ability of the full-precision model during quantization.
In this chapter, we first present the optimization objective of FQAT.
Next, we analyze the gradient conflict issue of quantization parameters caused by dual optimization objectives and propose a layer-wise freezing mechanism.
Finally, we introduce the disorder-guided adaptive freezing algorithm of FQAT.



\subsection{Optimization Objective of FQAT}
Following SAGM~\citep{wang2023sharpness}, we adopt three optimization objectives for flatness-oriented minimization over the training distributions $\mathcal{D}$: 
(a) empirical risk loss \( \mathcal{L}(\theta; \mathcal{D}) \), 
(b) perturbed loss \( \mathcal{L}_p(\theta; \mathcal{D}) \), and 
(c) the surrogate gap \( h(\theta) \) $\triangleq$ \( \mathcal{L}_{p}(\theta; \mathcal{D}) - \mathcal{L}(\theta; \mathcal{D}) \), 
where $\theta$ is the parameters of DNNs. 
Minimizing \( \mathcal{L}(\theta; \mathcal{D}) \) and \( \mathcal{L}_p(\theta; \mathcal{D}) \) finds low-loss regions, while minimizing \( h(\theta) \) ensures a flat minimum. This combination improves both training performance and OOD generalization. 
Hence, the overall optimization objective is:
\begin{equation}
\underset{\theta}\min \big(\mathcal{L}(\theta; \mathcal{D}), \mathcal{L}_p(\theta;\mathcal{D}),  h(\theta)\big)
\end{equation}
SAGM~\citep{wang2023sharpness} proposed the overall objective can  be achieved by the following formulation : 
\begin{equation}
\underset{\theta}\min \big(\mathcal{L}(\theta; \mathcal{D}) + \mathcal{L}_p(\theta - \alpha \nabla \mathcal{L}(\theta; \mathcal{D}); \mathcal{D}) \big), 
\end{equation}
where \( \alpha \) is the hyperparameter.
Then we incorporate QAT quantizer $Q(\theta;\mathbf{s})$ to the full-precison objective of flatness: 
\begin{small}
\begin{equation}
    \min_{\theta,\mathbf{s}} \ \mathcal{L} \left( Q(\theta; \mathbf{s}); \mathcal{D} \right) + \mathcal{L}_p \Big( Q \big( \theta - \alpha \nabla \mathcal{L} 
    \left(Q(\theta; \mathbf{s}); \mathcal{D} \right); \mathbf{s} \big); \mathcal{D} \Big),
\label{eq:sagm_qat}
\end{equation}
\end{small}

where $\mathbf{s}$ denote the trainable parameters (scaling factors) of quantizers. 
\begin{figure}[t]
    \centering
    \vspace{0cm}
    \includegraphics[width=\linewidth]{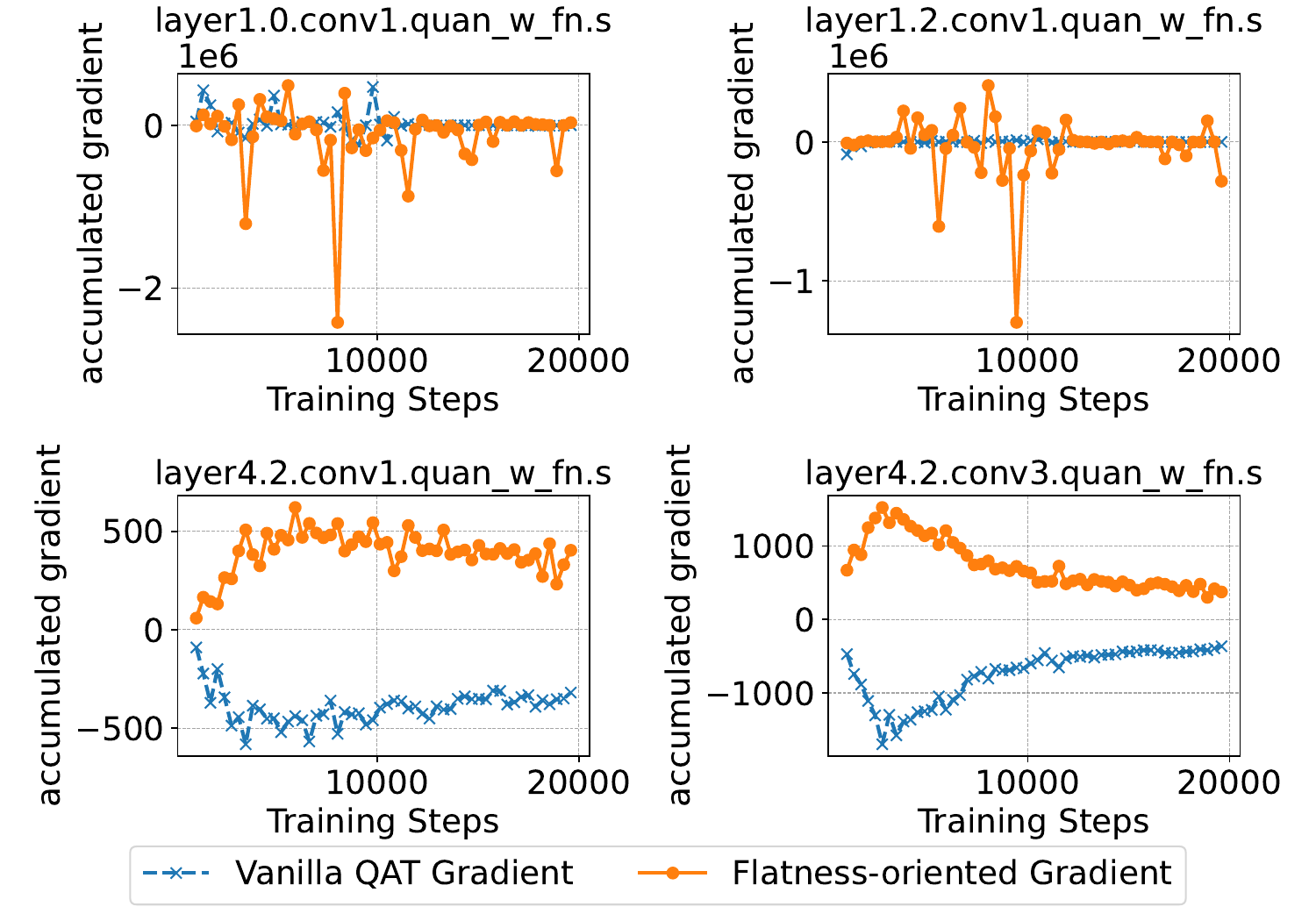}
    \caption{Results of cumulative gradients every 350 steps in the 4-bit test on the PACS ART domain, revealing conflicts of V-QAT gradient and flatness-oriented gradient.}
    \label{fig:conflict}
\end{figure}

\begin{figure}[t]
    \centering
    \includegraphics[width=\linewidth]{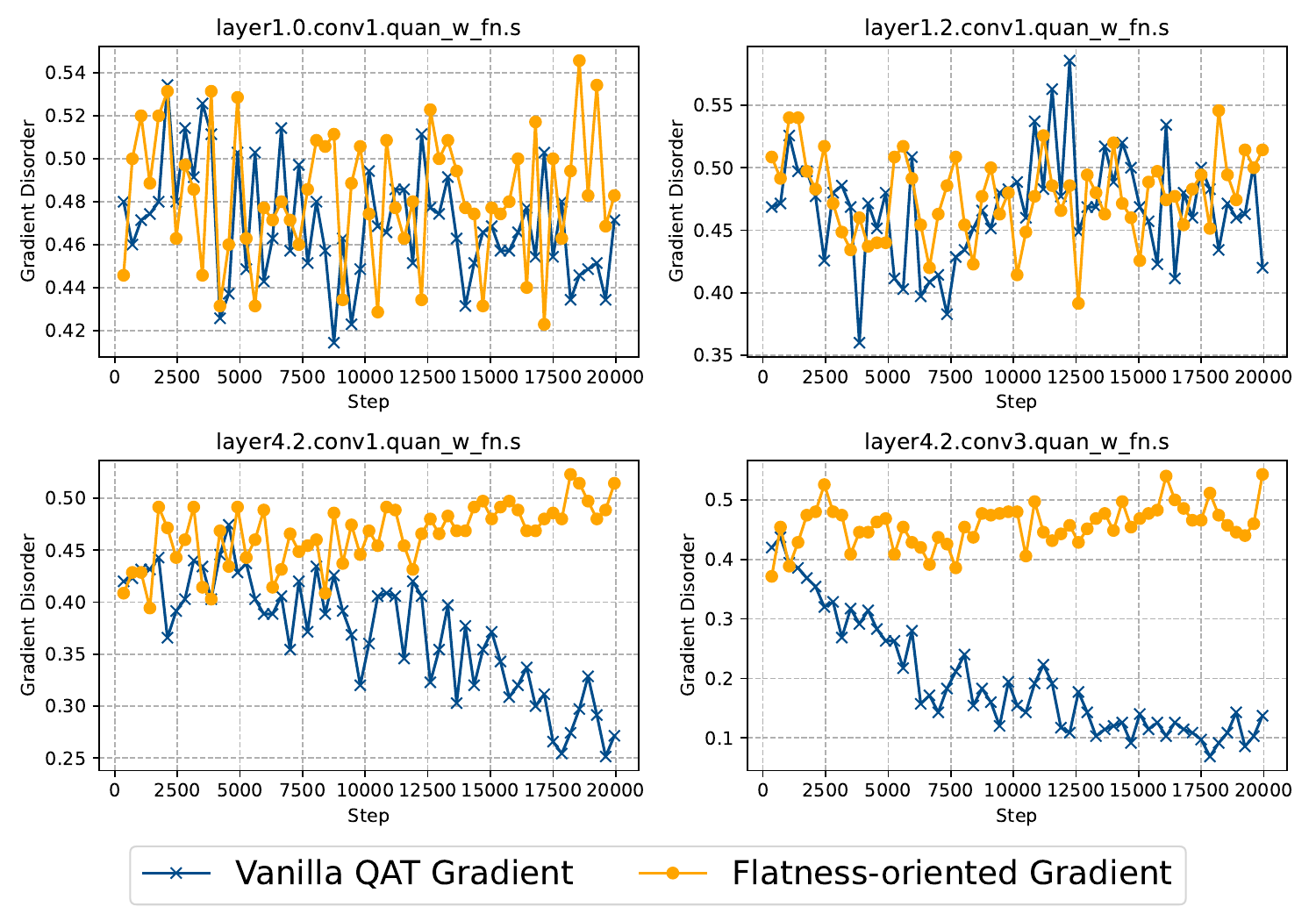}
    \caption{Results of V-QAT gradient and flatness-oriented gradient disorder of scale factors over 350 steps in the 4-bit test on the PACS ART domain, revealing in certain layers, the gradient disorder of V-QAT gradient decreases significantly as training progresses.}
    \label{fig:disorder}
\end{figure}

Specifically, during the optimization process, a gradient is obtained through the original QAT process. Based on this gradient, the original parameters are perturbed to obtain perturbed parameters. Then, a flat-related gradient is obtained by performing the QAT process with these perturbed parameters. By jointly update original parameters by the original quantization gradient and the flat-related gradient, a flatter loss landscape with low loss value is achieved.
More details about the optimization objective are provided in supplementary materials.


\subsection{Layer-wise Freezing Mechanism for Gradient Conflict of Quantizer}
As demonstrated by the results of \textit{LSQ+SAGM} and \textit{EWGS+SAGM} in Table \ref{tab:combined_avg},
the OOD performance improve in certain cases (e.g., 5bit on PACS, 4, 5, 8bit on DomainNet), 
though directly adopting the objective.
This phenomenon 
suggests the potential for enhancing quantized
model performance from a flatness perspective.
However, in some cases (for example, 3, 4bit on PACS), performance degrades significantly. 
We hypothesize that the improper optimization of the scale factor may be the cause of this performance degradation. 
The scale factor, used to portray the characteristic of weight and activation distribution~\citep{tang2022mixed},
is highly sensitive to the perturbations~\citep{esser2019learned,liu2023oscillation}.
Optimization of quantization parameters (i.e., the scale factor in the quantizer) is of importance for OOD generalization.
\begin{table}[t]

\caption{Results of 4-bit quantization tests with perturbed scale factors on the Clipart and Infograph domains (DomainNet). A / B indicates OOD test accuracy of Clipart and Infograph. Original OOD Performance is 60.21 / 15.81. The notation x\% denotes that the scale factor is multiplied by x\%.  \textcolor{red}{↓}: degradation; \textcolor{green}{↑}: improvement; —: no change. Proved that the scale factors has not fully converged, and the sensitivity of different layers varies}
\label{tab:performance_comparison}
\centering
\setlength{\tabcolsep}{2.5pt}
\renewcommand{\arraystretch}{1}
\resizebox{0.48\textwidth}{!}{
\begin{tabular}{lcccc}
\hline
\textbf{Layer} & \textbf{80\%} & \textbf{90\%} & \textbf{110\%} & \textbf{120\%} \\
\hline
L3.0.c1.w.s  & 
60.30\textcolor{green}{\tiny$\uparrow$} / 15.93\textcolor{green}{\tiny$\uparrow$} & 
60.15\textcolor{red}{\tiny$\downarrow$} / 15.94\textcolor{green}{\tiny$\uparrow$} & 
59.96\textcolor{red}{\tiny$\downarrow$} / 15.62\textcolor{red}{\tiny$\downarrow$} & 
59.82\textcolor{red}{\tiny$\downarrow$} / 15.38\textcolor{red}{\tiny$\downarrow$} \\[1pt]

L3.0.c1.a.s  & 
60.47\textcolor{green}{\tiny$\uparrow$} / 16.12\textcolor{green}{\tiny$\uparrow$} & 
60.31\textcolor{green}{\tiny$\uparrow$} / 15.90\textcolor{green}{\tiny$\uparrow$} & 
60.10\textcolor{red}{\tiny$\downarrow$} / 15.72\textcolor{red}{\tiny$\downarrow$} & 
59.93\textcolor{red}{\tiny$\downarrow$} / 15.65\textcolor{red}{\tiny$\downarrow$} \\[1pt]

L1.0.c1.w.s  & 
60.25\textcolor{green}{\tiny$\uparrow$} / 15.60\textcolor{red}{\tiny$\downarrow$} & 
60.14\textcolor{red}{\tiny$\downarrow$} / 15.61\textcolor{red}{\tiny$\downarrow$} & 
60.32\textcolor{green}{\tiny$\uparrow$} / 15.48\textcolor{red}{\tiny$\downarrow$} & 
60.18\textcolor{red}{\tiny$\downarrow$} / 15.27\textcolor{red}{\tiny$\downarrow$} \\[1pt]

L1.0.c1.a.s  & 
60.23\textcolor{green}{\tiny$\uparrow$} / \textcolor{black}{15.81\tiny—} &  
60.22\textcolor{green}{\tiny$\uparrow$} / 15.85\textcolor{green}{\tiny$\uparrow$} & 
60.26\textcolor{green}{\tiny$\uparrow$} / 15.78\textcolor{red}{\tiny$\downarrow$} & 
60.24\textcolor{green}{\tiny$\uparrow$} / 15.67\textcolor{red}{\tiny$\downarrow$} \\
\hline
\end{tabular}}

\end{table}

Compared to full-precision training, Eq.~(\ref{eq:sagm_qat}) introduces several scale factors \( \mathbf{s}\)
in two optimization objective functions,
thus generating two sets of gradients.  
One set originates from the optimization of vanilla QAT, denoted as \( \mathbf{g}_{\text{va}} \) and derived from \( \mathcal{L}(\cdot) \).  
The other set is the newly introduced flatness-oriented gradient, aimed at enhancing generalization ability and denoted as \( \mathbf{g}_{\text{flat}} \), derived from \( \mathcal{L}_{p}(\cdot) \). 

To investigate the interaction between these two gradients, we visualized their sum during the training process. As shown at the top of Figure \ref{fig:conflict}, the flatness-oriented gradient (yellow) exhibits significantly higher volatility and more outliers compared to the vanilla QAT gradient (blue). Moreover, for certain layers, \( \mathbf{g}_{\text{va}} \) and \( \mathbf{g}_{\text{flat}} \) are in opposite directions and tend to cancel each other out (bottom of Figure \ref{fig:conflict}).
Here we define this phenomenon as the \textit{sub-optimal equilibrium state}. This conflict between \( \mathbf{g}_{\text{va}} \) and \( \mathbf{g}_{\text{flat}} \) prevents the scale factors from fully converging, which can substantially degrade the performance of QAT in OOD scenarios.

To analyze gradient behavior from the perspective of directional fluctuation, we define the gradient disorder as a metric to quantify the degree of directional variability during training.

\begin{definition}
\textbf{Gradient Disorder:} 
For every \(K\) steps of training, we define two gradient sequences: \textcolor{black}{ \(G_1 = \{\mathbf{g}_1, \mathbf{g}_2, \dots, \mathbf{g}_{K-1}\}\) and \(G_2 = \{\mathbf{g}_2, \mathbf{g}_3, \dots, \mathbf{g}_K\}\)}, where \(\mathbf{g}_j\) denotes the gradient at step \(j\). 
Let \(\operatorname{sgn}(\cdot)\) denote the element-wise sign function. The \textit{gradient disorder} is defined as:
\textcolor{black}{\begin{equation}
\delta = \frac{1}{K-1} \sum_{i=1}^{K-1} \mathbbm{1} \left( \operatorname{sgn}(\mathbf{G}_1^{(i)}) \neq \operatorname{sgn}(\mathbf{G}_2^{(i)}) \right),
\end{equation}}
where $\mathbbm{1}(\cdot)$ is the indicator function. 
$\delta$ measures the proportion of adjacent gradients with opposite directions in the gradient sequence, indicating the degree of directional fluctuation.
\end{definition}

Figure~\ref{fig:disorder} shows that in some layers, the gradient disorder of \( \mathbf{g}_{\text{va}} \) decreases significantly during training, implying increasingly consistent gradient directions—a somewhat counterintuitive result. In contrast, \( \mathbf{g}_{\text{flat}} \) gradient disorder remains high across layers. Layers with lower \( \mathbf{g}_{\text{va}} \) disorder (bottom of Figure~\ref{fig:disorder}) display opposite and similar-magnitude gradients in Figure~\ref{fig:conflict}, suggesting a tendency to settle into a \textit{sub-optimal equilibrium state}.
\begin{figure}[t]
    \centering
    \vspace{0.3cm}
    \includegraphics[width=\linewidth]{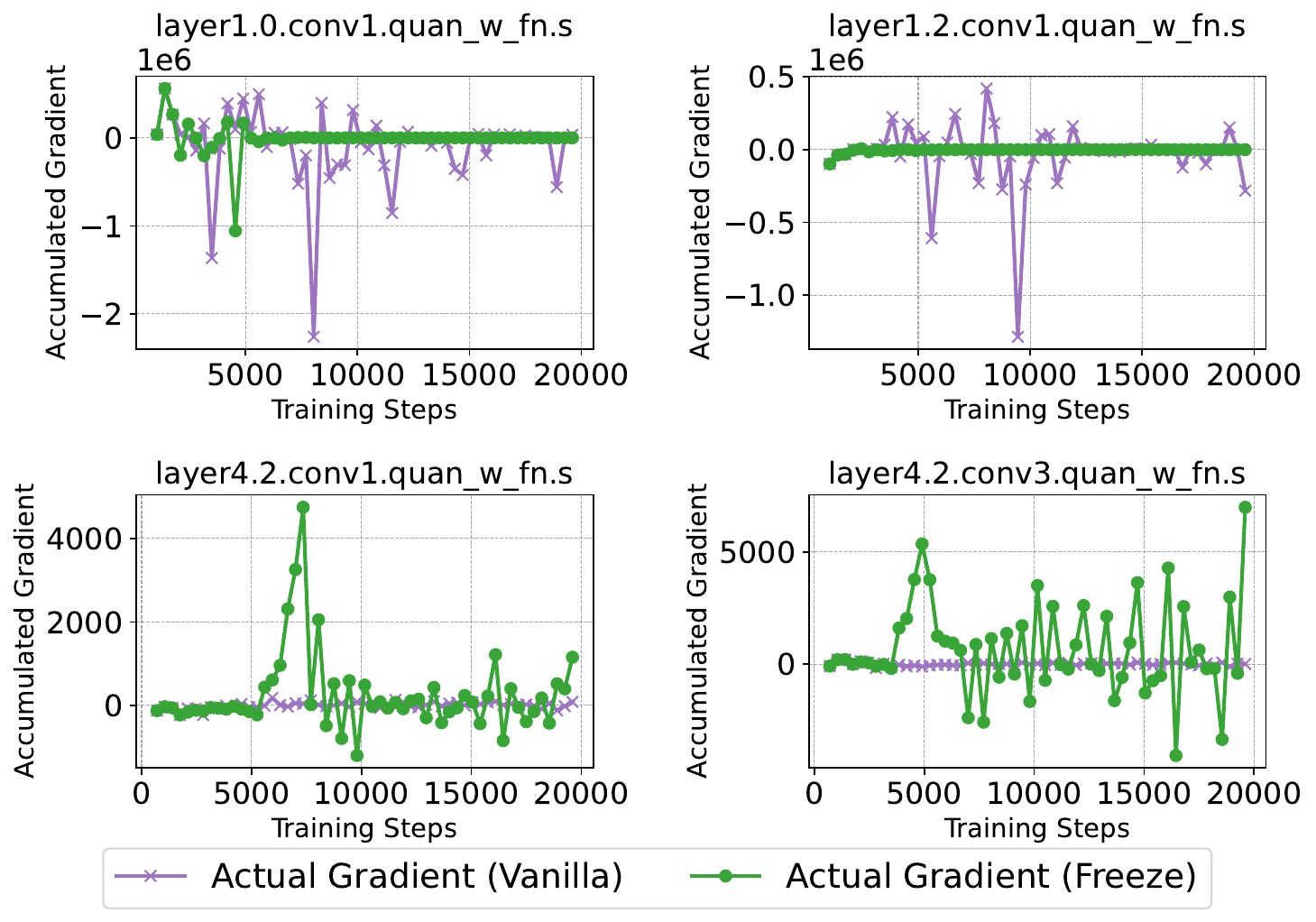}
    \caption{Results of the 4-bit quantization test on the PACS ART domain demonstrate the impact of layer freezing over 350 training steps, with V-QAT gradient disorder serving as an indicator. The results show that freezing specific layers (the bottom layers) strengthens the training of their actual gradients (only flatness-oriented). Meanwhile, the actual gradients (V-QAT and flatness-oriented) of the unfrozen layers become more stable, with fewer outliers. This demonstrates that V-QAT gradient disorder effectively regulates gradient fluctuations across different layers, providing a reliable metric for optimizing layer-wise training dynamics.}
    \label{fig:vs}
\end{figure}

\begin{assumption}
\textbf{Impact of Scale Factors with Low \(\mathbf{g}_{{\text{va}}}\) Gradient Disorder on Training:} 
Scale factors with overly low \(\mathbf{g}_{{\text{va}}}\) gradient disorder hinder their \(\mathbf{g}_{{\text{flat}}}\) training, causing insufficient convergence and indirectly affecting other scale factors convergence.
\end{assumption}

To verify this, we draw inspiration from weight freezing strategies ~\citep{liu2023oscillation,tang2024retraining,nagel2022overcoming} and conduct an experiment using the gradient disorder of 
\(\mathbf{g}_{{\text{va}}}\)	
  as a freezing indicator. Specifically, we freeze the \(\mathbf{g}_{{\text{va}}}\) of scale factors once they begin to exhibit low gradient disorder, allowing only \(\mathbf{g}_{{\text{flat}}}\) to update the parameters until the end of training.

Figure~\ref{fig:conflict} shows the gradients of \(\mathbf{g}_{{\text{va}}}\) and \(\mathbf{g}_{{\text{flat}}}\) without freezing. In the top layers, \(\mathbf{g}_{{\text{flat}}}\) exhibits large fluctuations and outliers, while in the bottom layers, \(\mathbf{g}_{{\text{flat}}}\) and \(\mathbf{g}_{{\text{va}}}\) reach \textit{sub-optimal equilibrium state}.
Figure~\ref{fig:disorder} demonstrates that after some training, the gradient disorder of \(\mathbf{g}_{{\text{va}}}\) decreases in the bottom layers. Thus, our freezing strategy freezes \(\mathbf{g}_{{\text{va}}}\) in these layers after a certain number of steps, using only \(\mathbf{g}_{{\text{flat}}}\) for updates. 
Note that the original update actual gradient is \(\mathbf{g}_{{\text{va}}} + \mathbf{g}_{{\text{flat}}}\), but after freezing, the update gradient consists solely of \(\mathbf{g}_{{\text{flat}}}\). 
Figure~\ref{fig:vs} compares the actual gradients with and without freezing, showing that the gradients of the upper layers (high disorder, unfreeze) stabilize significantly after other layers freezing. In contrast, in the bottom layers (low disorder, freeze), after a certain number of steps, the actual gradient only involves \(\mathbf{g}_{{\text{flat}}}\) and continues to evolve, indicating continued training rather than insufficient convergence due to the cancellation of \(\mathbf{g}_{{\text{va}}}\). A more detailed explanation is as follows: regarding the dynamics of $g_{\text{flat}}$, once freezing is applied (e.g., to layer4.2.conv1 at 5k training steps), $g_{\text{flat}}$ exhibits two distinct phases: (1) an initial phase of unidirectional updates between approximately 5k and 7.5k steps, confirming the suppressive influence of $g_{\text{va}}$; and (2) a subsequent phase of balanced oscillations from 7.5k to 20k steps, indicating that the model is undergoing stable convergence, as shown in Fig~\ref{fig:vs}.
This confirms the hypothesis and validates that gradient disorder can serve as an effective indicator for guiding which layers to freeze, helping alleviate training instability.

\subsection{Disorder-guided
Adaptive Freezing Algorithm}

To prevent suboptimal convergence resulting from the persistent freezing of \(\mathbf{g}_{{\text{va}}}\) in certain layers without timely unfreezing, we design an adaptive freezing strategy:
for every \(K\) steps, we assess the gradient disorder \(\delta_{t,\mathbf{s}_i}\) of \(\mathbf{g}_{{\text{va}}}\) for each scale factor \(\mathbf{s}_i\), calculated from the \(\mathbf{g}_{{\text{va}}}\) sequence stored over the previous \(K\) steps. If \(\delta_{t,\mathbf{s}_i}\) falls below the threshold \(r\), we freeze \(\mathbf{g}_{{\text{va}}}\) of \(\mathbf{s}_i\) for the next \(K\) steps, updating only with \(\mathbf{g}_{{\text{flat}}}\). Otherwise, we unfreeze \(\mathbf{g}_{{\text{va}}}\), updating with \(\mathbf{g}_{{\text{va}}} + \mathbf{g}_{{\text{flat}}}\). The complete procedure is detailed in supplementary materials.

This adaptive selective freezing strategy effectively enhances model performance in OOD scenarios by adaptively managing gradient updates, while avoiding convergence issues due to excessive freezing.

%% file: experiment.tex
\section{Experiment}
\label{sec:experiment}
\subsection{Experimental Setup and Implementation Details}



\begin{figure}
    \centering
    \includegraphics[width=\linewidth]{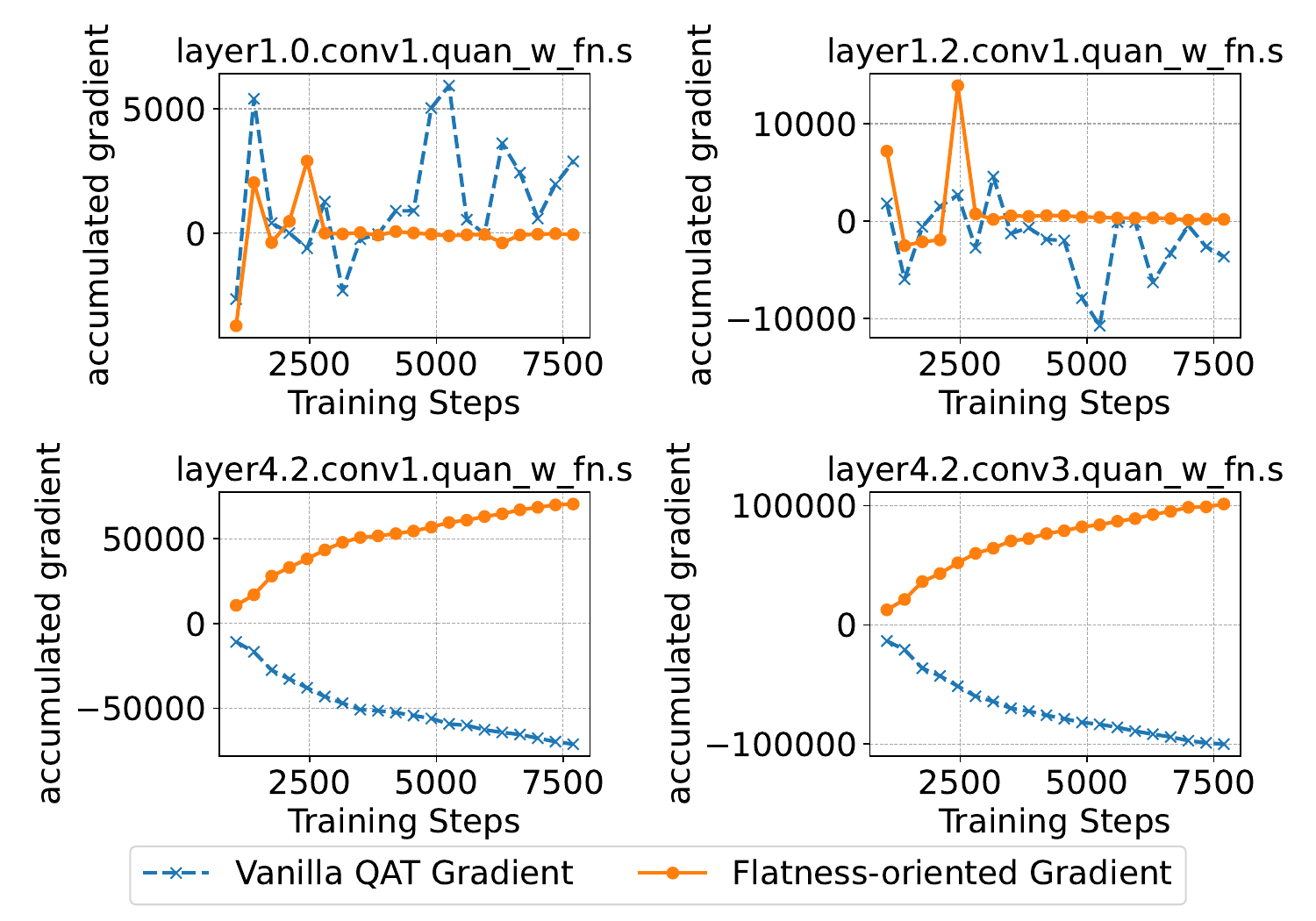}
    \caption{Results of cumulative gradients every 2111 steps in the 3-bit test on the DomainNet Clipart and Infograph domainsrevealing few anomalous gradients, with V-QAT gradient dominating.}
    \label{fig:3bitdomain}
    \vspace{-0.5cm}
\end{figure}

\textbf{Datasets and evaluation protocol.} We conduct a comprehensive evaluation on three widely used OOD datasets: PACS~\citep{Li_2017_ICCV}, containing 9,991 images across 7 categories and 4 domains; OfficeHome~\citep{venkateswara2017deep} with 15,588 images spanning 65 categories and 4 domains; DomainNet~\citep{peng2019domain}, the largest benchmark comprising 586,575 images across 345 categories and 6 domains.
We basically follow the evaluation protocol of DomainBed~\citep{gulrajani2020search}, including the optimizer, data split, and model selection, where we adopt train-domain validation as our model selection strategy for all algorithms in our experiments. 
For PACS and OfficeHome, for each time we treat one domain as the test domain and other domains as training domains, which is the leave-one-domain-out protocol commonly adopted in OOD evaluation. 
For DomainNet, following \citet{yu2024rethinking}, we divide the domains into three groups: (1) \textit{Clipart} and \textit{Infograph}, (2) \textit{Painting} and \textit{Quickdraw}, and (3) \textit{Real} and \textit{Sketch}. Then we employ the leave-one-group-out protocol, where we treat one group of two domains as test domains and other two groups as training domains each time. 
For the number of training steps, for full-precision models we set it as 5,000 for PACS and OfficeHome, 15,000 for DomainNet following \citet{cha2021swad}, while for quantization training we use 20,000 for PACS and OfficeHome, 50,000 for DomainNet. 
To reduce time cost, for quantization training we conduct validation and testing for DomainNet only after 45,000 steps. Each performance is reported as the average of two runs with seeds 0 and 23, while the ablation study is conducted with a single seed (seed 0) to reduce computational costs.

\noindent
\textbf{Quantization.} We follow established practices in QAT literature by employing the QAT method LSQ~\citep{esser2019learned} and EWGS~\citep{lee2021network} to quantize both weights and activations. The quantization scale factors are learned with a fixed learning rate of \(1 \times 10^{-5}\). We use Mean Squared Error (MSE) range estimation~\citep{nagel2021whitepaperneuralnetwork} to determine the quantization parameters for weights and activations. Due to the risk of test data information leakage of supervised pretrained weights revealed by \citet{yu2024rethinking}, we employ MoCo-v2~\citep{chen2020improvedbaselinesmomentumcontrastive} pretrained ResNet-50 as initialization as recommended. Then we fine-tune the model using Empirical Risk Minimization (ERM) to obtain a full-precision model with generalization capabilities, which serves as the baseline for quantization. The weights and activations are fully quantized, except for the first convolutional layer, which quantizes only the activations, and the final linear layer, which remains unquantized, striking a balance between efficiency and model capacity. For DomainNet, we apply LSQ and EWGS to quantize model to 4, 5, and 8bit precision, as 3bit quantization yields minimal conflicts (Figure~\ref{fig:3bitdomain}). For PACS, we use LSQ with 3, 4, and 5bit quantization. For OfficeHome, both LSQ and EWGS are applied at 3bit precision.

\subsection{Hyperparameter settings}
Given the substantial computational resources required by the original DomainBed setup, we adjust the hyperparameter search space and conduct grid search to reduce computational cost following SAGM~\citep{wang2023sharpness}. 
The search space of learning rate is \{1e-5, 3e-5, 5e-5\}, and the dropout rate is fixed as zero. 
The batch size of each training domain is set as $32$ for PACS and OfficeHome, $24$ for DomainNet. 
Following SAM~\citep{foret2020sharpness}, we fix the hyperparameter $\rho=0.05$. 
Following SAGM~\citep{wang2023sharpness}, we set $\alpha$ in \Cref{eq:sagm_qat} as $0.001$ for PACS and OfficeHome, $0.0005$ for DomainNet, and set weight decay as 1e-4 for PACS and OfficeHome, 1e-6 for DomainNet. 
For PACS and OfficeHome, the gradient disorder threshold \(r\) is selected from \(\{0.28, 0.30, 0.32\}\) for 3-bit, 4-bit and 5-bit quantization. The number of freeze steps is selected from \(\{300, 350, 400\}\) for both 4-bit and 5-bit quantization, and from \(\{100, 150, 200\}\) for 3-bit quantization. 
For DomainNet, \(r\) is selected from \(\{0.20, 0.25\}\) for both 4-bit and 5-bit quantization, and from \(\{0.3, 0.35\}\) for 8-bit quantization. The number of freeze steps is chosen from \(\{3000, 4000\}\) for both 4-bit and 5-bit quantization, and from \(\{1500, 2000\}\) for 8-bit quantization. We determine the hyperparameter space based on single-domain observation and apply it to other test domains, validating the robustness of our method.
To reduce the high computational cost, we first select the shared hyperparameters, i.e. learning rate, weight decay, through grid search, which serve as the base hyperparameter configuration. Then we fix the base configuration and conduct further grid search on our specific hyperparameters, i.e. freeze steps, freeze threshold.

\subsection{Main Results}

We conducted a comprehensive evaluation of our method on the PACS, OfficeHome and DomainNet datasets across various quantization bit-widths (see Table~\ref{tab:combined_avg}), highlighting three key advantages: 
\textbf{(i) significant improvements in I.D and OOD performance}, 
\textbf{(ii) greater enhancement in OOD performance compared to I.D performance}, and 
\textbf{(iii) improved training stability}. Across all quantization bit-widths, our method achieved the best performance in both validation and test sets on the three datasets under the quantization schemes. On the DomainNet dataset, the I.D performance surpassed the full-precision performance at most bits, with our method achieving an OOD test accuracy of 41.27\% (LSQ) and 41.01\% (EWGS) at 8-bit quantization, exceeding the full-precision accuracy of 40.95\% and setting a new state-of-the-art result. At 4-bit and 5-bit quantization, the OOD performance also approached full-precision levels. In terms of OOD performance preservation, despite the baseline accuracy of the validation set being significantly higher than that of the test set, the test accuracy gains across various bit-widths on the DomainNet dataset (compared to directly introducing SAGM) remained close to the corresponding validation set gains. On the PACS dataset, the test accuracy gains consistently exceeded the validation set gains compare to directly introducing SAGM (3-bit: +5.24\% test vs +2.61\% validation ; 4-bit: +3.07\% test vs +2.37\% validation; 5-bit: +1.72\% test vs +1.54\% validation), further confirming the significant enhancement of our method's generalization capability beyond I.D optimization. For OfficeHome dataset, both LSQ and EWGS exhibit significant OOD performance degradation at low-bit quantization, while our method achieves substantial gains. In terms of training stability, our method focuses on resolving gradient conflicts. Experimental results show that, in most cases, the standard deviations of our method's I.D and OOD performance are smaller than those of directly introducing SAGM, validating the effectiveness of our gradient coordination mechanism. More experimental results are provided in supplementary materials.

\subsection{Ablation Study}
In our analysis, we validated the effectiveness of freezing \(\mathbf{g}_{{\text{va}}}\) when gradient disorder falls below a specific threshold, coupled with periodically reselecting the freeze set to stabilize quantization training in the OOD scenario. This naturally leads to the question: what would occur if these strategies were modified? For instance, what if scale factors with gradient disorder above the threshold were frozen instead, or both gradients (\(\mathbf{g}_{{\text{va}}}\) and \(\mathbf{g}_{{\text{flat}}}\)) were frozen simultaneously, or unfreezing were avoided after the initial freeze? Furthermore, what would happen if a decoupled alternating update approach were adopted, where one step updates \(\mathbf{g}_{{\text{va}}}\) and the next updates \(\mathbf{g}_{{\text{flat}}}\)? Exploring these variations would offer deeper insights into the effectiveness of each component of our method.

Table~\ref{tab:ablation} highlights the importance of our proposed modules. The \textit{w/o Unfreeze} strategy shows a significant performance drop on PACS, emphasizing the necessity of our phased freezing strategy. The \textit{Reverse Freeze} strategy also results in notable accuracy degradation, validating the effectiveness of our freezing metric. Other strategies further demonstrate the superiority of our gradient selection methodology, confirming the robustness and effectiveness of our approach.

\begin{figure}
    \centering
    \vspace{-0.2cm}
\includegraphics[width=\linewidth]{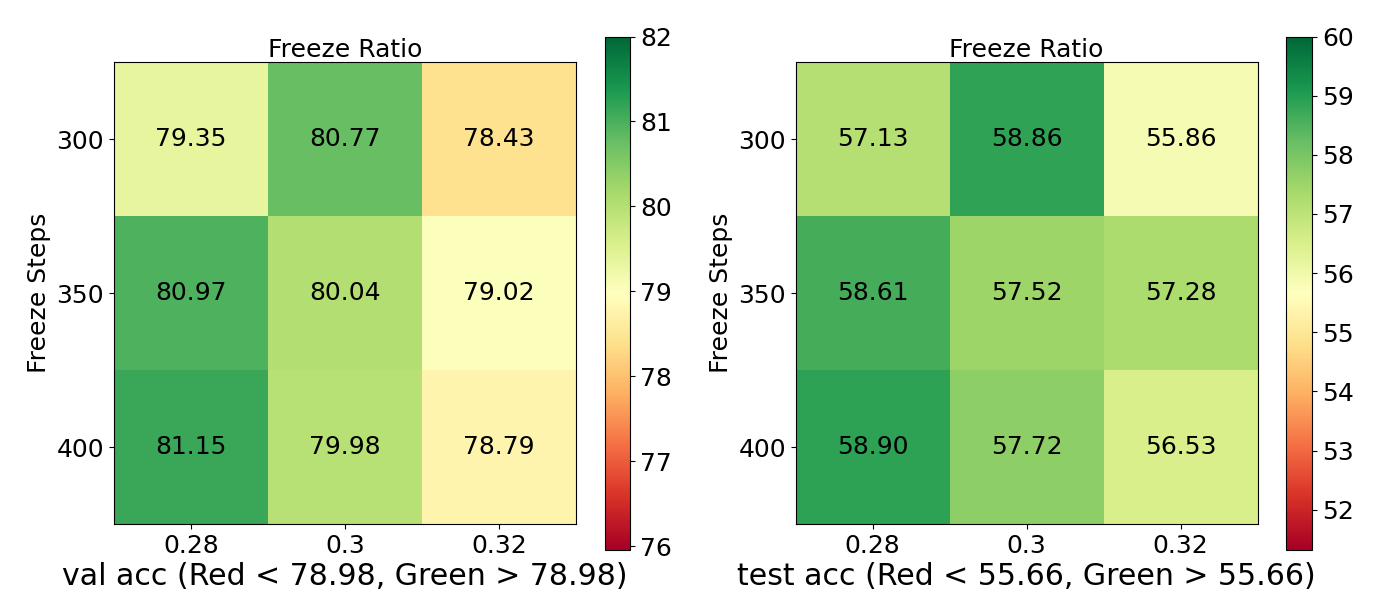}
    \caption{4-bit \textbf{LSQ} PACS (seed 0, 23): Val acc (I.D mean) and test acc (OOD mean) were evaluated via grid search on freeze steps and freeze ratio. Each cell shows mean results for all domains with the corresponding hyperparameters. Green: improvement over directly introducing flat objective; red: degradation. Baseline: Val 78.98 / Test 55.66. Most configurations show improvements, proving our method’s robustness to hyperparameters within a reasonable range.} 
    \label{fig:hyparam}
\end{figure}
\begin{figure*}[htbp]
    \centering
    
        \begin{minipage}[b]{0.05\textwidth}
            \centering
            \raisebox{0.5\height}{\rotatebox{90}{\textbf{Original}}}  
        \end{minipage}
        \begin{minipage}[b]{0.18\textwidth}
            \centering
            \includegraphics[width=\linewidth]{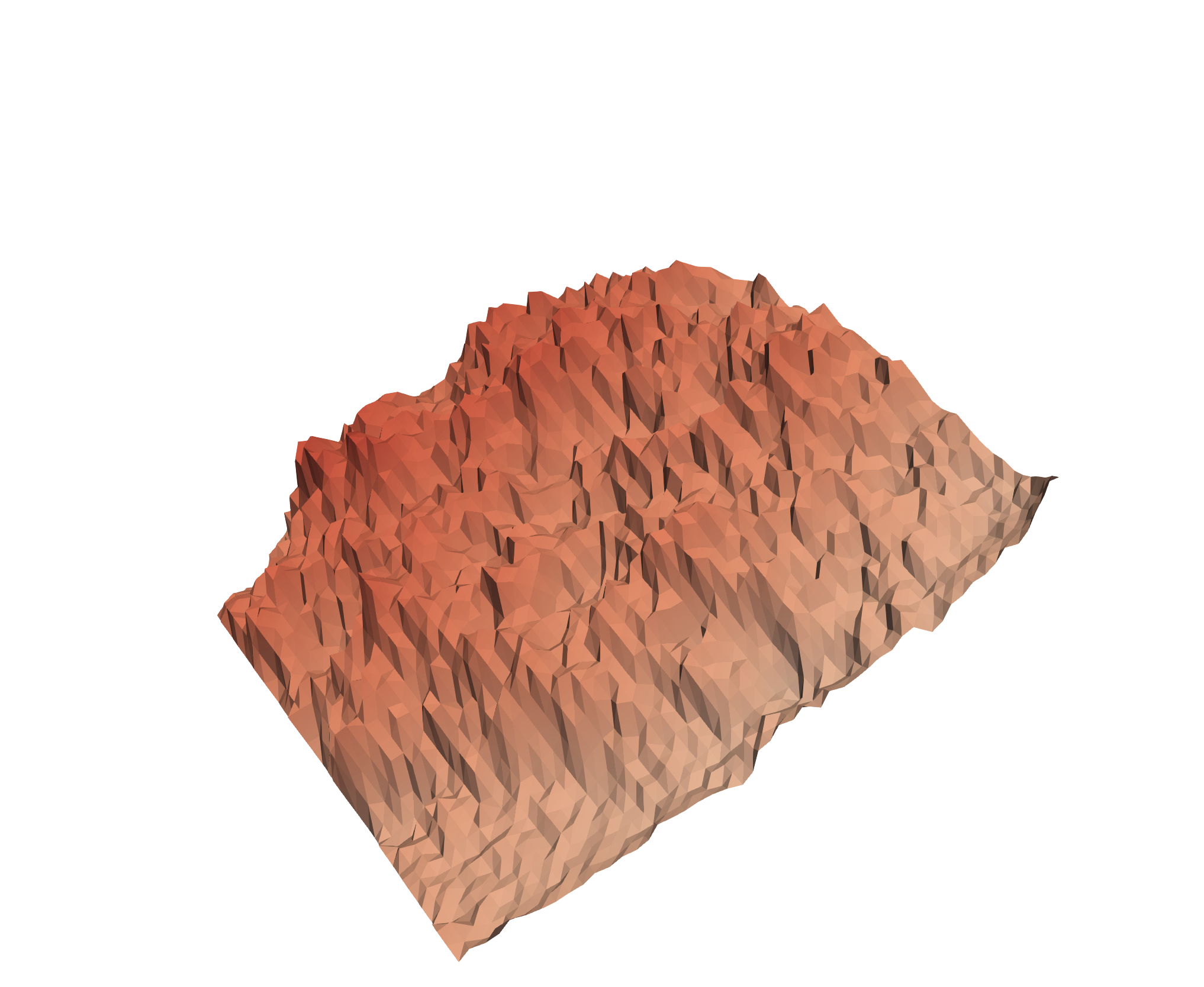}
            \label{fig:bad0}
        \end{minipage}
        \hspace{0.02\textwidth}  
        \begin{minipage}[b]{0.18\textwidth}
            \centering
            \includegraphics[width=\linewidth]{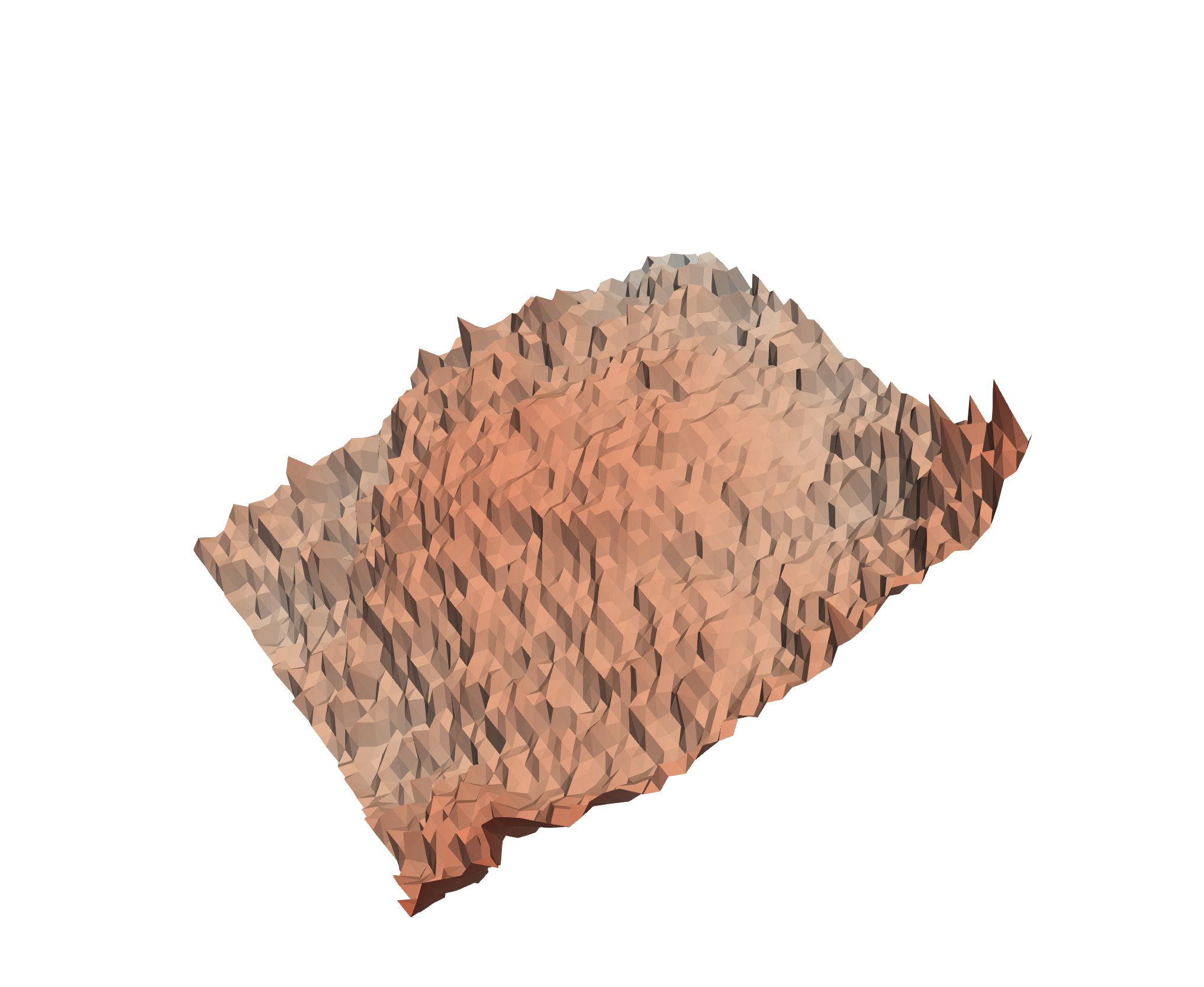}
            \label{fig:bad1}
        \end{minipage}
        \hspace{0.02\textwidth}
        \begin{minipage}[b]{0.18\textwidth}
            \centering
            \includegraphics[width=\linewidth]{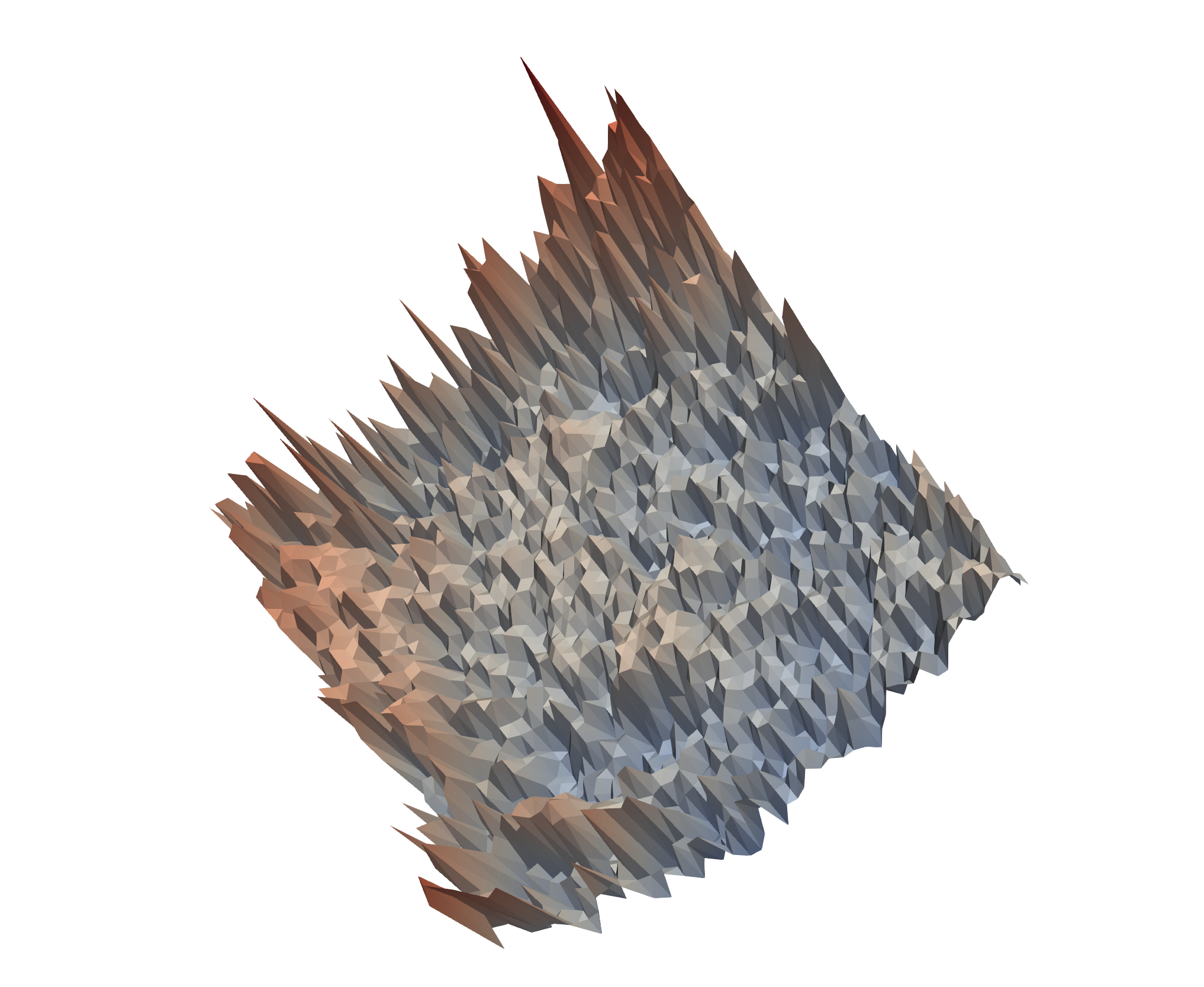}
            \label{fig:bad3}
        \end{minipage}
        \hspace{0.02\textwidth}
        \begin{minipage}[b]{0.18\textwidth}
            \centering
            \includegraphics[width=\linewidth]{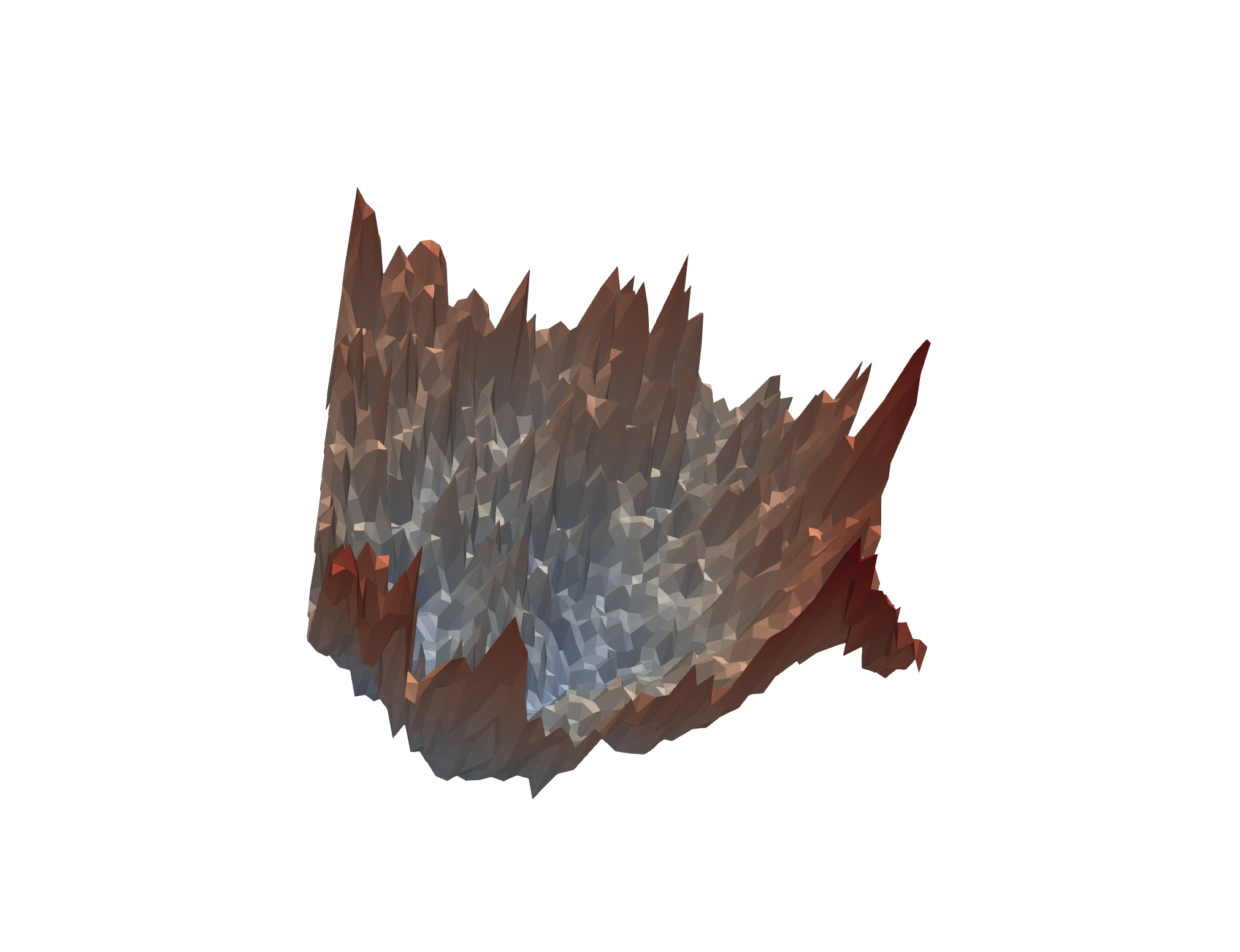}
            \label{fig:origin}
        \end{minipage}
        \vspace{-1em}  
        
        \begin{minipage}[b]{0.05\textwidth}
            \centering
            \raisebox{1.5\height}{\rotatebox{90}{\textbf{Ours}}}  
        \end{minipage}
        \begin{minipage}[b]{0.18\textwidth}
            \centering
            \includegraphics[width=\linewidth]{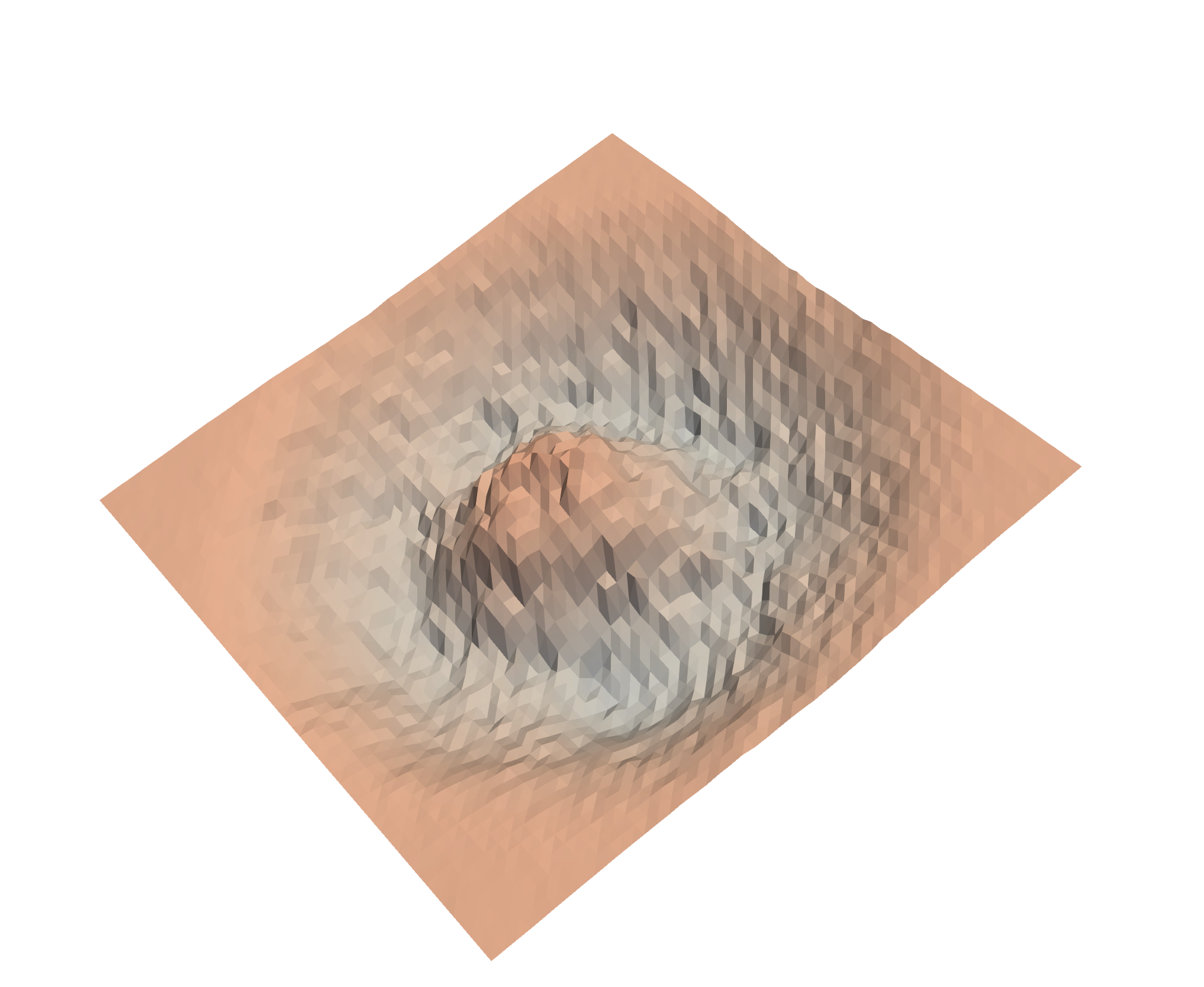}
            \caption*{\textbf{Art}}
            \label{fig:good0}
        \end{minipage}
        \hspace{0.02\textwidth}
        \begin{minipage}[b]{0.18\textwidth}
            \centering
            \includegraphics[width=\linewidth]{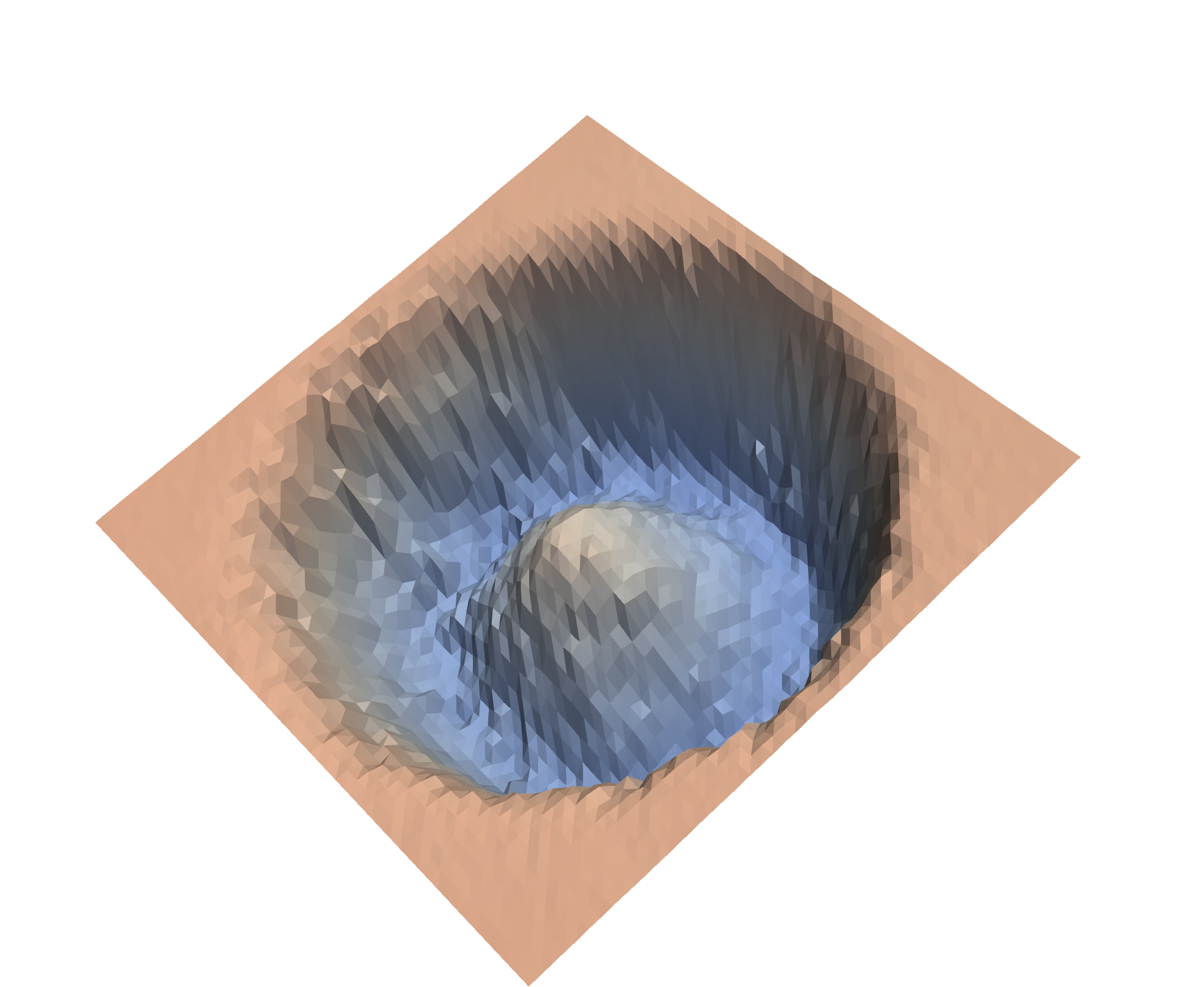}
            \caption*{\textbf{Cartoon}}
            \label{fig:good1}
        \end{minipage}
        \hspace{0.02\textwidth}
        \begin{minipage}[b]{0.18\textwidth}
            \centering
            \includegraphics[width=\linewidth]{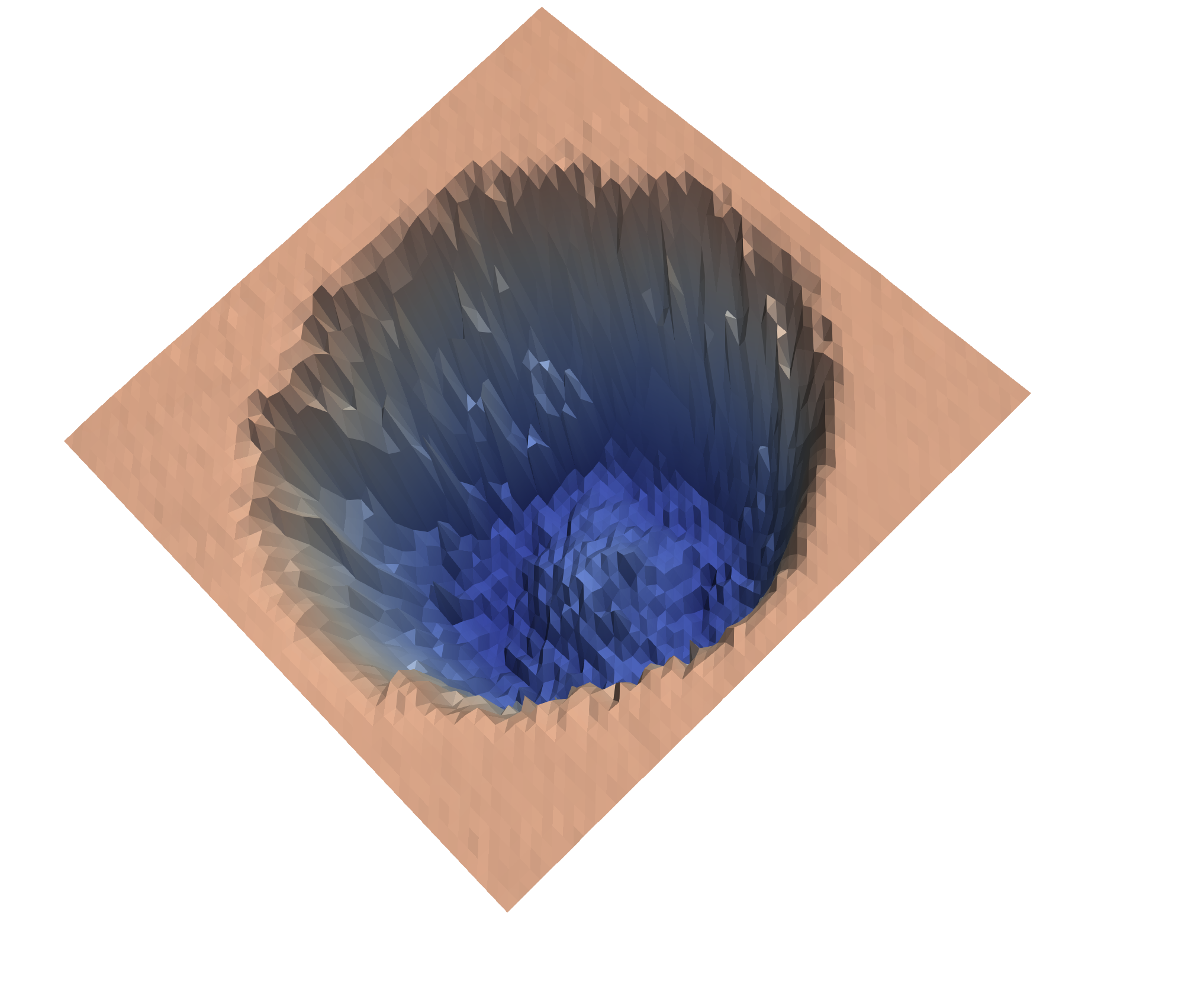}
            \caption*{\textbf{Photo}}
            \label{fig:good3}
        \end{minipage}
        \hspace{0.02\textwidth}
        \begin{minipage}[b]{0.18\textwidth}
            \centering
            \includegraphics[width=\linewidth]{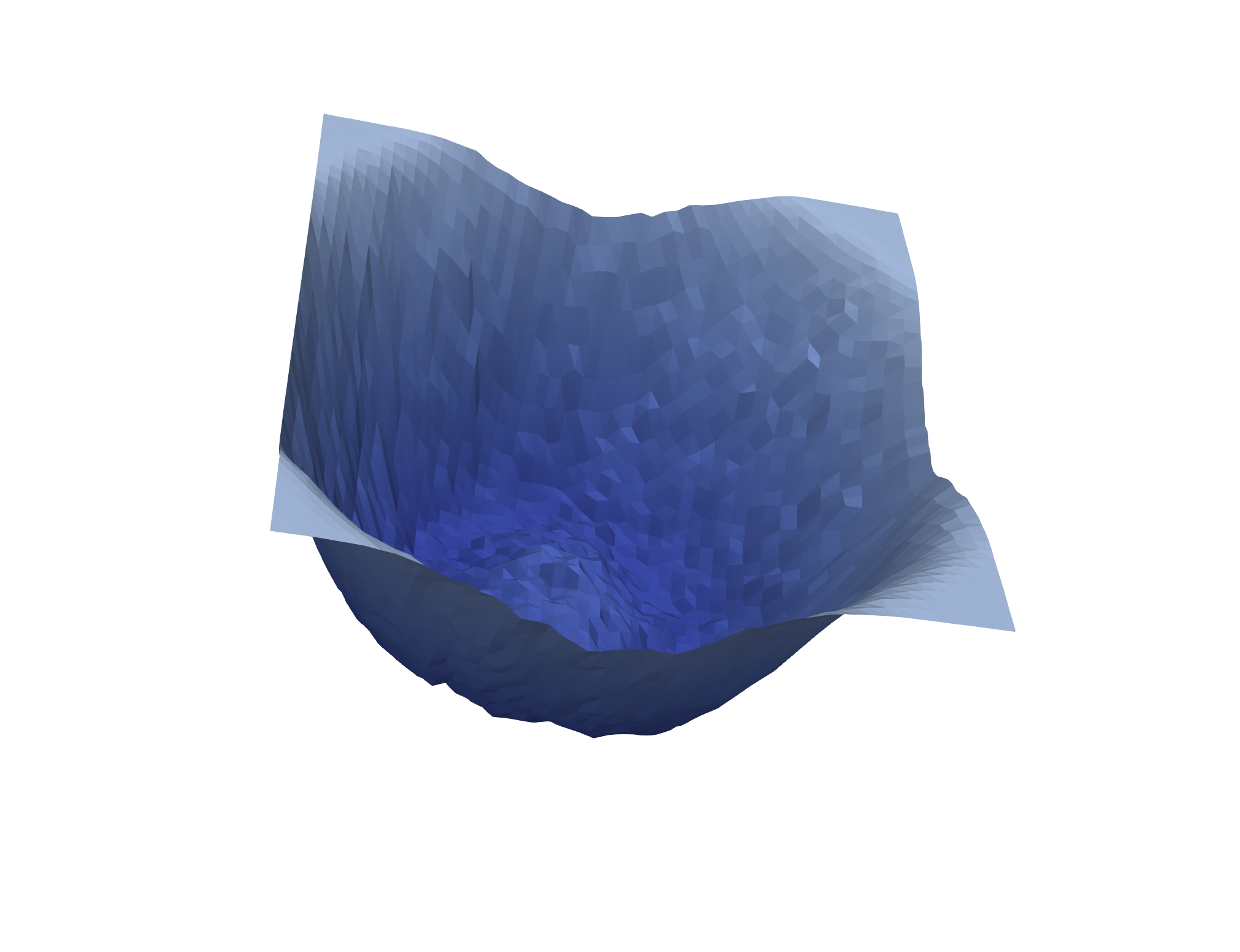}
            \caption*{\textbf{Sketch}}
            \label{fig:smooth}
        \end{minipage}

    \captionsetup{justification=raggedright,singlelinecheck=false} 
    \caption{\textcolor{black}{Visualization of the loss landscape across different domains (PACS 4bit), where the categories from left to right are Art, Cartoon, Photo, and Sketch. Blue indicates low loss values, while red represents high loss values. The top row displays the results of LSQ + SAGM, and the bottom row shows the results using our proposed method. Our approach consistently achieves flatter loss surfaces with lower loss values across all four domains in PACS, demonstrating its effectiveness in optimizing the loss landscape.}}
    \label{fig:bad_good_comparison}
    \vspace{-0.5cm}
\end{figure*}

\begin{table}[h!]
\centering
\caption{Ablation Study: \textbf{LSQ} experiments on seed 0 evaluate Val (I.D validation mean accuracy) and Test (OOD test mean accuracy). Compared variants include: Ours (proposed FQAT), Alter Update (Alternate V-QAT and flatness-oriented gradients updates), Freeze Both (freeze QAT and flatness-oriented gradients), w/o Unfreeze (no unfreezing after initial freeze), and Reverse Freeze (freeze above the threshold, unlike the original). Best performance is in bold, validating the effectiveness of our design.}
\label{tab:ablation}
\scriptsize 
\renewcommand{\arraystretch}{0.6} 
\setlength{\tabcolsep}{2pt} 
\resizebox{0.48\textwidth}{!}{ 
\begin{tabular}{@{}llccc@{}}
\toprule
\rowcolor{lightgray} Dataset & Method & Bit-width (W/A) & Val & Test \\ 
\midrule
\multirow{5}{*}{DomainNet} 
& Ours & 4/4 & 62.45 & \textbf{40.60} \\ 
& Alter Update & 4/4 & 57.41 & 37.15 \\ 
& Freeze Both & 4/4 & 61.03 & 39.36 \\ 
& w/o Unfreeze & 4/4 & \textbf{62.50} & 40.53 \\ 
& Reverse Freeze & 4/4 & 59.32 & 38.24 \\ 
\midrule
\multirow{5}{*}{PACS} 
& Ours & 4/4 & \textbf{81.20} & \textbf{59.33} \\ 
& Alter Update & 4/4 & 80.66 &  58.36\\ 
& Freeze Both & 4/4 & 79.74 & 56.98 \\ 
& w/o Unfreeze & 4/4 & 79.50 & 57.20 \\ 
& Reverse Freeze & 4/4 & 80.16 & 56.89 \\ 
\bottomrule
\end{tabular}
}
\end{table}

\subsection{Hyperparameter Sensitivity Analysis}

Figure \ref{fig:hyparam} demonstrates the robustness of our method within a reasonable hyperparameter search space. We evaluated the performance of our approach on PACS (4-bit) using fixed hyperparameters across all domains (note that grid-searching for optimal hyperparameters per domain would yield better results). The results show that our method outperforms the baseline in both validation and test accuracy in most cases, indicating its insensitivity to hyperparameter variations. This suggests that simply defining a reasonable search range is sufficient for practical use.

During the experiments, we randomly selected one domain to observe the gradient disorder of 
\( \mathbf{g}_{{\text{va}}} \)
 under different freeze steps. Based on empirical observations, we determined the search space for freeze steps and freeze ratio, which was then directly applied to other domains without further tuning.

\subsection{\textcolor{black}{Loss Surface Visualization}}
Following the approach in~\citep{li2018visualizing}, Figure \ref{fig:bad_good_comparison} compares the loss surface visualizations across the four domains of PACS (4-bit LSQ) when using SAGM directly versus our proposed method. The results demonstrate that our method consistently achieves flatter loss surfaces with lower loss values across all domains, highlighting its effectiveness in optimizing the loss landscape. We also visualize loss surface of DomainNet (Clipart \& Infograph domains) in supplementary materials. LSQ + SAGM shows a smoother loss surface compared to PACS, which further explains why our approach is more significantly effective on PACS than on DomainNet.

%% file: related_work.tex
\section{Related Work}
\subsection{Quantizaion-aware Training on CV} 
Quantizaion-aware training (QAT) involves inserting simulated quantization nodes and retraining the model, which achieves a better balance between accuracy and compression ratio~\citep{hubara2021accurate,pmlr-v119-nagel20a}. 
DoReFa~\citep{zhou2016dorefa} and PACT~\citep{choi2018pact} use low-precision weights and activations during the forward pass and utilize STE techniques~\citep{bengio2013estimating} during backpropagation to estimate gradients of the piece-wise quantization functions. LSQ~\citep{esser2019learned} adjusts the quantization function by introducing learnable step size scaling factors. EWGS~\citep{lee2021network} scales the gradients to precisely adjust the quantization error component in the gradients, reducing gradient conflicts caused by quantization errors. SAQ~\citep{liu2021sharpness} demonstrates that quantized models make the loss surface sharper. Recently, some works have explored the possibility of improving quantization performance by freezing unstable weights to further enhance results~\citep{nagel2022overcoming,tang2024retraining,liu2023oscillation}; however, these methods have only considered the identically distributed (I.D) data. 
Due to distribution shifts in test data—which often occur in practical applications—the quality and reliability of quantized models cannot be guaranteed~\citep{hu2022characterizing}. 
\subsection{OOD Generalization on CV}
In practical applications, when deploying machine learning models, test data distribution may differ from the training distribution, a common phenomenon known as distribution shift~\citep{liu2021towards,yu2024survey,koh2021wilds}. 
OOD generalization methods aim to enhance a model's ability to generalize to OOD distributions~\citep{wang2022generalizing, zhou2022domain}. Common strategies include domain alignment ~\citep{muandet2013domain, li2018domain,zhao2020domain}, meta learning~\citep{li2018learning,balaji2018metareg,dou2019domain}, data augmentation ~\citep{zhou2021domain,carlucci2019domain}, disentangled representation learning ~\citep{zhang2022towards} and utilization of causal relations~\citep{mahajan2021domain,lv2022causality}. 
Inspired by previous studies of flat minima~\citep{izmailov2018averaging,foret2020sharpness,liu2022towards,zhuang2022surrogate,zhang2023gradient}, flatness-aware methods start to gain attention and exhibit remarkable performance in OOD generalization ~\citep{cha2021swad,wang2023sharpness,zhang2023flatness}, such as SAGM~\citep{wang2023sharpness}, improving generalization ability by optimizing the angle between weight gradients to reduce gradient conflicts. 
As far as we know, existing works all focus on improving the OOD generalization ability of full-precision models, while neglecting the generalization issues of quantized models. Our work is the first to fill this gap by identifying the issue and offering a solution.


%% file: CONCLUSION_AND_FUTURE_WORK.tex
\section{Conclusion and Future Work}
Our work validates a crucial yet overlooked problem: quantization harms the OOD gereralization performance of full-precision CV models. 
Further, we proposes FQAT, a flatness-oriented QAT method, incorporating a layer-wise freezing mechanism to mitigate gradient conflicts and a disorder-guided adaptive freezing algorithm for dynamic layer adjustment. 
The extensive
experiments on several OOD benchmarks demonstrate
the superiority of the proposed method over state-of-the-art
baselines.

\textbf{Limitations and future work.}
Due to limitations in experimental resources, we did not explore alternative flatness objectives to further investigate their effects on scale factor gradients. Additionally, we recommend evaluating the robustness of our approach in conjunction with various full-precision OOD generalization techniques. Our experiments also revealed that different domains exhibit varying sensitivities to the scale factor, suggesting that a deeper investigation into the relationship between domain characteristics and scale factor dynamics could offer promising directions for future optimization. Finally, our current analysis is primarily conducted under the ReLU activation setting, where all QAT zero points are aligned at zero. Extending the analysis to other activation functions and quantization settings remains an important avenue for future work.

%% file: appendix.tex
\appendix
\onecolumn

\section{Optimization Objective Explanation}
\label{sec:A-op}
The overall objective of SAGM can be achieved by the following formulation:
\[
\underset{\theta}\min \big(\mathcal{L}(\theta; \mathcal{D}) + \mathcal{L}_p(\theta - \alpha \nabla_\theta \mathcal{L}(\theta; \mathcal{D}); \mathcal{D}) \big),
\]
where \(\alpha\) is a hyperparameter. The second term can be further rewritten as in SAGM:
\[ \min_{\theta} 
\mathcal{L}\left(\theta + \hat{\epsilon} - \alpha \nabla_\theta \mathcal{L}(\theta; \mathcal{D}); \mathcal{D}\right),
\]
where
\[
\hat{\epsilon} = \rho \frac{\nabla_\theta \mathcal{L}(\theta; \mathcal{D})}{\|\nabla_\theta \mathcal{L}(\theta; \mathcal{D})\|},
\]
 \(\rho\) is a hyperparameter. The process begins with backpropagation based on the loss function \(\mathcal{L}(\theta; \mathcal{D})\) to obtain the gradient \(\nabla_\theta \mathcal{L}(\theta; \mathcal{D})\). The perturbation magnitude $\hat{\epsilon}$ is then calculated to perturb the original weights \(\theta\) to obtain new weights \(\theta_p = \theta + \hat{\epsilon}\). Next, the SAGM target perturbed weights are computed as:
\[
\theta_{\text{SAGM}} = \theta_p - \alpha \nabla_\theta \mathcal{L}(\theta; \mathcal{D}).
\]
Using these new weights, the loss is recalculated as \(\mathcal{L}\left(\theta_{\text{SAGM}}; \mathcal{D}\right)\), yielding a new gradient \(\nabla_{\theta_{\text{SAGM}}} \mathcal{L}(\theta_{\text{SAGM}}; \mathcal{D})\). Finally, the original weights \(\theta\) are updated using both \(\nabla_{\theta_{\text{SAGM}}} \mathcal{L}(\theta_{\text{SAGM}}; \mathcal{D})\) and \(\nabla_\theta \mathcal{L}(\theta; \mathcal{D})\), specifically by averaging the two gradients during implementation.

When directly integrating the SAGM objective into our quantization process, the key difference lies in the computation of gradients. The first gradient is obtained using quantized weights \(Q(\theta, s)\) through the Straight-Through Estimator, i.e., \(\nabla_\theta \mathcal{L}(Q(\theta, s); \mathcal{D})\). The perturbation magnitude is then calculated as:
\[
\hat{\epsilon} = \rho \frac{\nabla_\theta \mathcal{L}(Q(\theta, \mathbf{s}); \mathcal{D})}{\|\nabla_\theta \mathcal{L}(Q(\theta, \mathbf{s}); \mathcal{D})\|},
\]
resulting in \(\theta' = \theta + \hat{\epsilon}\). For the second gradient, the actual computation uses the quantized perturbed weights \(Q(\theta', \mathbf{s})\), yielding \(\nabla_{\theta'} \mathcal{L}(Q(\theta', \mathbf{s}); \mathcal{D})\). These two gradients are used to update \(\theta\), while the scale parameter \(\mathbf{s}\) remains unperturbed. During this process, \(\mathbf{s}\) also receives two gradients, \(\mathbf{g}_{\text{va}}\) = $\nabla_s \mathcal{L}(Q(\theta, \mathbf{s}); \mathcal{D})$and \(\mathbf{g}_{\text{flat}}\) = $\nabla_s \mathcal{L}(Q({\theta'}, \mathbf{s}); \mathcal{D})$, which are used to update \(\mathbf{s}\).

\section{pseudocode}
\label{sec:pseudocode}
\begin{figure}[htbp] 
    \centering
    \vspace{0cm}
    \begin{minipage}{0.45\textwidth} 
        \begin{algorithm}[H]
        \caption{\fontsize{10}{10}\selectfont Disorder-guided Adaptive Freezing Algorithm}
        \label{alg:freeze}
        \begin{algorithmic}[1]
        \fontsize{7}{7}\selectfont 
        \Require overall training steps \(T\), step interval \(K\), threshold \(r\) and scale factors \(\{\mathbf{s}_1, \mathbf{s}_2, \dots, \mathbf{s}_n\}\)
        \State Init \(t \gets 0\), \(\text{freeze}[\mathbf{s}_i] \gets False\) for all \(\mathbf{s}_i, i=1,\cdots,n\)
        \While{\(t < T\)}
            \For{$i=1,\cdots,n$}
                \If{\(\text{freeze}[\mathbf{s}_i]\)} 
                    \State Update \(\mathbf{s}_i\) with \( \mathbf{g}_{{\text{flat}}} \)
                \Else
                    \State Update \(\mathbf{s}_i\) with \( \mathbf{g}_{{\text{va}}}, \mathbf{g}_{{\text{flat}}} \)
                \EndIf
            \EndFor
            \If{\(t \bmod K = 0\)}
                \For{$i=1,\cdots,n$}
                    \State Compute \(\delta_{t,\mathbf{s}_i}\)
                     \If{\(\delta_{t,\mathbf{s}_i} < r\)}
                 \State     \text{freeze}[$\mathbf{s}_i$] $\gets True$
                     \Else
                   \State     \text{freeze}[$\mathbf{s}_i$] $\gets False$  
                \EndIf
                \EndFor
            \EndIf
            \State \(t \gets t + 1\)
        \EndWhile
        \end{algorithmic}
        \end{algorithm}
    \end{minipage}
\end{figure}

\section{More experimental results}
\label{sec:C-ex}

\begin{figure}
    \centering
    \vspace{0cm}
\includegraphics[width=0.3\linewidth]{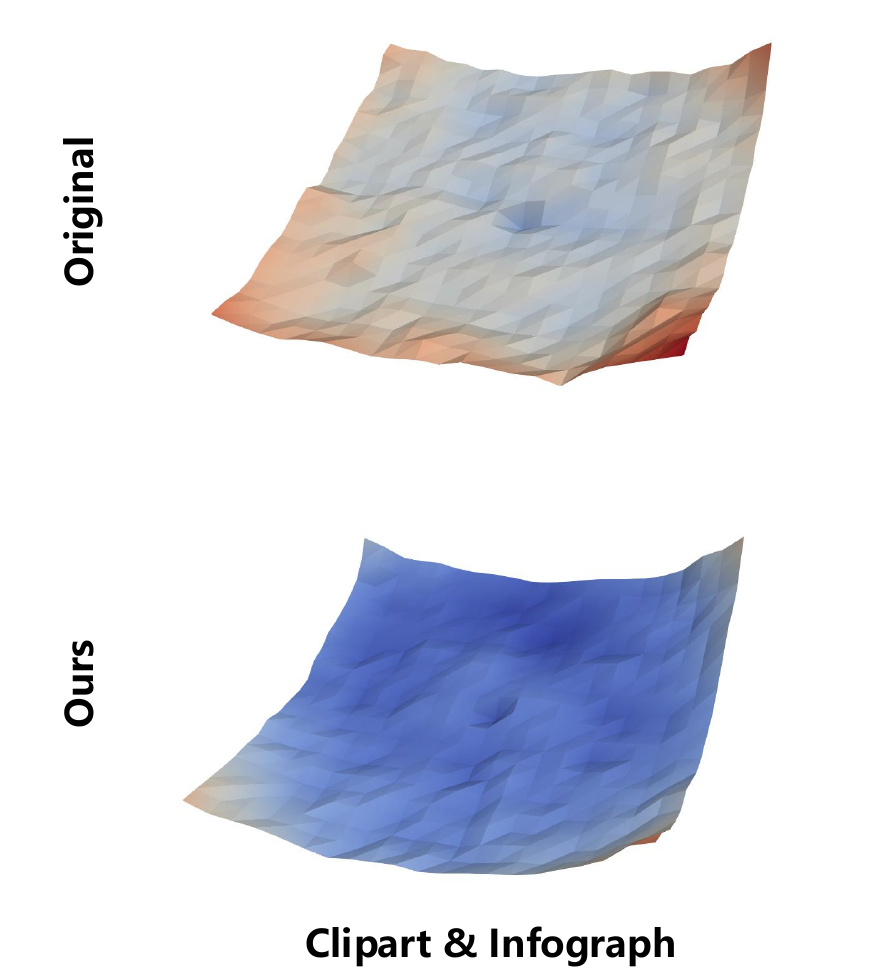}
    \caption{Visualization of the loss landscape of DomainNet Clipart \& Infograph domains (4bit). The top row displays the results of LSQ + SAGM, and the bottom row shows the results using our proposed method. } 
    \label{fig:loss2}
   \vspace{1cm}
\end{figure}

\begin{table}[htbp]
  \centering
  \caption{4-bit LSQ Results on DomainNet Validation Set (val)}

  \begin{tabular}{lccccccc}
    \toprule
    Method & Clipart & Infograph & Painting & Quickdraw & Real & Sketch & Avg \\
    \midrule
    LSQ & 65.35±0.98 & 65.35±0.98 & 59.71±0.15 & 59.71±0.15 & 57.80±0.02 & 57.80±0.02 & 60.95±0.38 \\
    LSQ + SAGM & 66.13±0.36 & 66.13±0.36 & 61.15±0.05 & 61.15±0.05 & 58.09±0.37 & 58.09±0.37 & 61.79±0.26 \\
    Ours & 67.19±0.01 & 67.19±0.01 & 61.83±0.68 & 61.83±0.68 & 58.54±0.05 & 58.54±0.05 & 62.52±0.25 \\
    \bottomrule
  \end{tabular}
\end{table}

\begin{table}[htbp]
  \centering
  \caption{4-bit LSQ Results on DomainNet Test Set (test)}
  \begin{tabular}{lccccccc}
    \toprule
    Method & Clipart & Infograph & Painting & Quickdraw & Real & Sketch & Avg \\
    \midrule
    LSQ & 59.86±0.59 & 15.21±0.44 & 44.81±0.13 & 14.74±0.01 & 52.51±0.19 & 47.72±0.10 & 39.14±0.24 \\
    LSQ + SAGM & 60.95±0.22 & 15.73±0.09 & 46.75±0.08 & 16.09±0.20 & 52.23±0.39 & 48.69±0.38 & 40.07±0.23 \\
    Ours & 61.13±0.13 & 16.13±0.01 & 47.02±0.79 & 16.43±0.01 & 53.47±0.06 & 49.43±0.14 & 40.6±0.19 \\
    \bottomrule
  \end{tabular}
\end{table}

\begin{table}[htbp]
  \centering
  \caption{5-bit LSQ Results on DomainNet Validation Set (val)}
  \begin{tabular}{lccccccc}
    \toprule
    Method & Clipart & Infograph & Painting & Quickdraw & Real & Sketch & Avg \\
    \midrule
    LSQ & 65.22±0.62 & 65.22±0.62 & 59.64±0.23 & 59.64±0.23 & 56.44±0.27 & 56.44±0.27 & 60.43±0.37 \\
    LSQ + SAGM & 67.36±0.51 & 67.36±0.51 & 61.86±0.31 & 61.86±0.31 & 58.97±0.44 & 58.97±0.44 & 62.73±0.42 \\
    Ours & 67.60±0.52 & 67.60±0.52 & 62.24±0.28 & 62.24±0.28 & 59.74±0.28 & 59.74±0.28 & 63.19±0.36 \\
    \bottomrule
  \end{tabular}
\end{table}

\begin{table}[htbp]
  \centering
  \caption{5-bit LSQ Results on DomainNet Test Set (test)}
  \begin{tabular}{lccccccc}
    \toprule
    Method & Clipart & Infograph & Painting & Quickdraw & Real & Sketch & Avg \\
    \midrule
    LSQ & 60.06±0.26 & 15.51±0.20 & 45.14±0.14 & 14.30±0.03 & 49.72±0.30 & 46.05±0.03 & 38.46±0.16 \\
    LSQ + SAGM & 61.41±0.19 & 16.40±0.23 & 47.76±0.51 & 15.71±0.08 & 52.73±0.68 & 49.03±0.53 & 40.51±0.37 \\
    Ours & 61.77±0.26 & 16.31±0.42 & 47.93±0.23 & 15.83±0.01 & 53.91±0.45 & 49.82±0.16 & 40.93±0.25 \\
    \bottomrule
  \end{tabular}
\end{table}

\begin{table}[htbp]
  \centering
  \caption{8-bit LSQ Results on DomainNet Validation Set (val)}
  \begin{tabular}{lccccccc}
    \toprule
    Method & Clipart & Infograph & Painting & Quickdraw & Real & Sketch & Avg \\
    \midrule
    LSQ & 66.21±0.54 & 66.21±0.54 & 59.96±0.11 & 59.96±0.11 & 56.74±0.27 & 56.74±0.27 & 60.97±0.31 \\
    LSQ + SAGM & 68.60±0.09 & 68.60±0.09 & 62.38±0.29 & 62.38±0.29 & 58.64±0.71 & 58.64±0.71 & 63.21±0.36 \\
    Ours & 68.40±0.15 & 68.40±0.15 & 62.40±0.37 & 62.40±0.37 & 59.37±0.11 & 59.37±0.11 & 63.39±0.21 \\
    \bottomrule
  \end{tabular}
\end{table}

\begin{table}[htbp]
  \centering
  \caption{8-bit LSQ Results on DomainNet Test Set (test)}
  \begin{tabular}{lccccccc}
    \toprule
    Method & Clipart & Infograph & Painting & Quickdraw & Real & Sketch & Avg \\
    \midrule
    LSQ & 60.18±0.11 & 16.02±0.30 & 46.19±0.06 & 13.96±0.01 & 51.22±0.25 & 47.05±0.21 & 39.1±0.16 \\
    LSQ + SAGM & 62.49±0.13 & 17.45±0.07 & 48.22±0.34 & 15.26±0.04 & 53.80±1.02 & 49.37±0.51 & 41.1±0.35 \\
    Ours & 62.23±0.17 & 17.24±0.10 & 48.33±0.30 & 15.40±0.18 & 54.54±0.21 & 49.90±0.13 & 41.27±0.18 \\
    \bottomrule
  \end{tabular}
\end{table}

\begin{table}[htbp]
  \centering
  \caption{3-bit Quantization Results on PACS Validation Set (val)}
  \begin{tabular}{lccccc}
    \toprule
    Method & Art & Cartoon & Photo & Sketch & Avg \\
    \midrule
    LSQ & 81.64±0.52 & 73.47±1.51 & 79.72±0.52 & 74.72±0.16 & 77.39±0.68 \\
    LSQ + SAGM & 82.44±1.03 & 70.07±2.68 & 76.01±1.79 & 71.40±0.47 & 74.98±1.49 \\
    Ours & 82.81±0.20 & 74.23±1.54 & 78.62±1.78 & 74.68±1.30 & 77.59±1.21 \\
    \bottomrule
  \end{tabular}
\end{table}

\begin{table}[htbp]
  \centering
  \caption{3-bit Quantization Results on PACS Test Set (test)}
  \begin{tabular}{lccccc}
    \toprule
    Method & Art & Cartoon & Photo & Sketch & Avg \\
    \midrule
    LSQ & 39.26±0.82 & 54.74±3.94 & 61.23±1.95 & 62.18±2.23 & 54.35±2.23 \\
    LSQ + SAGM & 40.88±2.68 & 47.28±1.44 & 59.09±0.94 & 55.14±0.43 & 50.6±1.37 \\
    Ours & 43.14±0.43 & 55.46±3.60 & 63.14±1.38 & 61.63±0.49 & 55.84±1.48 \\
    \bottomrule
  \end{tabular}
\end{table}

\begin{table}[htbp]
  \centering
  \caption{4-bit Quantization Results on PACS Validation Set (val)}
  \begin{tabular}{lccccc}
    \toprule
    Method & Art & Cartoon & Photo & Sketch & Avg \\
    \midrule
    LSQ & 86.57±1.71 & 78.30±0.43 & 81.61±0.80 & 75.28±0.32 & 80.44±0.82 \\
    LSQ + SAGM & 85.18±1.02 & 77.05±3.11 & 80.19±1.60 & 73.50±0.11 & 78.98±1.46 \\
    Ours & 85.93±0.82 & 80.08±1.65 & 81.98±1.81 & 77.04±0.21 & 81.26±1.12 \\
    \bottomrule
  \end{tabular}
\end{table}

\begin{table}[htbp]
  \centering
  \caption{4-bit Quantization Results on PACS Test Set (test)}
  \begin{tabular}{lccccc}
    \toprule
    Method & Art & Cartoon & Photo & Sketch & Avg \\
    \midrule
    LSQ & 46.25±4.82 & 57.22±0.88 & 63.51±0.26 & 62.63±0.35 & 57.4±1.58 \\
    LSQ + SAGM & 45.97±0.52 & 53.94±2.72 & 62.57±2.10 & 60.16±1.35 & 55.66±1.67 \\
    Ours & 47.41±1.83 & 58.02±1.36 & 65.98±0.71 & 66.28±1.11 & 59.42±1.25 \\
    \bottomrule
  \end{tabular}
\end{table}

\begin{table}[htbp]
  \centering
  \caption{5-bit Quantization Results on PACS Validation Set (val)}
  \begin{tabular}{lccccc}
    \toprule
    Method & Art & Cartoon & Photo & Sketch & Avg \\
    \midrule
    LSQ & 84.53±0.04 & 80.89±2.93 & 83.81±1.03 & 71.05±0.21 & 80.07±1.05 \\
    LSQ + SAGM & 86.12±2.60 & 80.45±3.78 & 81.42±0.55 & 73.72±1.91 & 80.43±2.21 \\
    Ours & 87.49±1.76 & 81.03±3.39 & 83.30±1.14 & 76.04±0.26 & 81.97±1.64 \\
    \bottomrule
  \end{tabular}
\end{table}

\begin{table}[htbp]
  \centering
  \caption{5-bit Quantization Results on PACS Test Set (test)}
  \begin{tabular}{lccccc}
    \toprule
    Method & Art & Cartoon & Photo & Sketch & Avg \\
    \midrule
    LSQ & 43.29±1.01 & 58.50±3.33 & 67.37±0.97 & 60.37±0.00 & 57.38±1.33 \\
    LSQ + SAGM & 47.44±4.42 & 59.86±4.10 & 64.60±1.50 & 62.88±0.10 & 58.69±2.53 \\
    Ours & 49.60±3.36 & 61.51±4.05 & 66.65±1.83 & 63.87±0.48 & 60.41±2.43 \\
    \bottomrule
  \end{tabular}
\end{table}

\begin{table}[htbp]
  \centering
  \caption{8-bit EWGS Results on DomainNet Validation Set (val)}
  \begin{tabular}{lcccccc}
    \toprule
    Method & Clipart & Infograph & Painting & Quickdraw & Real & Sketch  \\
    \midrule
    EWGS & 66.20±0.28 & 66.20±0.28 & 59.79±0.16 & 59.79±0.16 & 57.12±0.06 & 57.12±0.06 \\
    EWGS + SAGM & 67.87±0.21 & 67.87±0.21 & 62.17±0.29 & 62.17±0.29 & 59.01±0.40 & 59.01±0.40 \\
    Ours & 67.82±0.21 & 67.82±0.21 & 62.31±0.28 & 62.31±0.28 & 59.20±0.04 & 59.20±0.04 \\
    \bottomrule
  \end{tabular}
\end{table}

\begin{table}[htbp]
  \centering
  \caption{8-bit EWGS Results on DomainNet Test Set (test)}
  \begin{tabular}{lcccccc}
    \toprule
    Method & Clipart & Infograph & Painting & Quickdraw & Real & Sketch  \\
    \midrule
    EWGS & 60.22±0.07 & 16.17±0.21 & 45.84±0.18 & 13.94±0.01 & 51.84±0.08 & 47.40±0.07 \\
    EWGS + SAGM & 62.09±0.07 & 16.76±0.12 & 48.16±0.31 & 15.51±0.18 & 53.89±0.27 & 49.13±0.33 \\
    Ours & 61.99±0.09 & 16.75±0.01 & 48.35±0.17 & 15.51±0.12 & 54.12±0.31 & 49.32±0.24 \\
    \bottomrule
  \end{tabular}
\end{table}

\begin{table}[htbp]
  \centering
  \caption{5-bit EWGS Results on DomainNet Validation Set (val)}
  \begin{tabular}{lcccccc}
    \toprule
    Method & Clipart & Infograph & Painting & Quickdraw & Real & Sketch \\
    \midrule
    EWGS & 66.32±0.03 & 66.32±0.03 & 60.20±0.12 & 60.20±0.12 & 57.11±0.32 & 57.11±0.32 \\
    EWGS + SAGM & 66.98±0.95 & 66.98±0.95 & 60.42±0.23 & 60.42±0.23 & 59.30±0.22 & 59.30±0.22 \\
    Ours & 67.32±1.04 & 67.32±1.04 & 61.18±0.16 & 61.18±0.16 & 59.93±0.17 & 59.93±0.17 \\
    \bottomrule
  \end{tabular}
\end{table}

\begin{table}[htbp]
  \centering
  \caption{5-bit EWGS Results on DomainNet Test Set (test)}
  \begin{tabular}{lcccccc}
    \toprule
    Method & Clipart & Infograph & Painting & Quickdraw & Real & Sketch \\
    \midrule
    EWGS & 60.20±0.12 & 15.73±0.01 & 46.15±0.32 & 14.14±0.22 & 51.02±0.54 & 46.88±0.25 \\
    EWGS + SAGM & 61.15±0.51 & 16.10±0.39 & 46.10±0.32 & 15.61±0.36 & 54.12±0.23 & 49.55±0.23 \\
    Ours & 61.50±0.49 & 16.13±0.47 & 46.75±0.26 & 15.85±0.08 & 55.04±0.14 & 49.98±0.14 \\
    \bottomrule
  \end{tabular}
\end{table}

\begin{table}[htbp]
  \centering
  \caption{4-bit EWGS Results on DomainNet Validation Set (val)}
  \begin{tabular}{lcccccc}
    \toprule
    Method & Clipart & Infograph & Painting & Quickdraw & Real & Sketch \\
    \midrule
    EWGS & 63.74±1.45 & 63.74±1.45 & 60.00±0.27 & 60.00±0.27 & 57.29±0.62 & 57.29±0.62 \\
    EWGS + SAGM & 64.44±0.19 & 64.44±0.19 & 60.12±0.22 & 60.12±0.22 & 58.79±0.49 & 58.79±0.49 \\
    Ours & 63.96±0.35 & 63.96±0.35 & 61.01±0.80 & 61.01±0.80 & 59.27±0.36 & 59.27±0.36 \\
    \bottomrule
  \end{tabular}
\end{table}

\begin{table}[htbp]
  \centering
  \caption{4-bit EWGS Results on DomainNet Test Set (test)}
  \begin{tabular}{lcccccc}
    \toprule
    Method & Clipart & Infograph & Painting & Quickdraw & Real & Sketch \\
    \midrule
    EWGS & 58.91±0.75 & 14.63±0.49 & 45.62±0.20 & 14.46±0.20 & 51.64±0.88 & 47.15±0.79 \\
    EWGS + SAGM & 59.82±0.11 & 15.71±0.21 & 45.63±0.40 & 16.04±0.01 & 53.52±0.50 & 49.18±0.40 \\
    Ours & 59.04±0.21 & 15.35±0.15 & 46.18±0.77 & 16.32±0.12 & 54.48±0.41 & 50.02±0.37 \\
    \bottomrule
  \end{tabular}
\end{table}

\begin{table}[htbp]
  \centering
  \caption{3-bit EWGS Results on OfficeHome Validation Set (val)}
  \begin{tabular}{lcccc}
    \toprule
    Method & Art & Clipart & Product & Real-World \\
    \midrule
    EWGS & 66.48±0.97 & 58.33±3.74 & 56.10±1.88 & 60.36±3.03 \\
    EWGS + SAGM & 62.83±3.25 & 58.81±0.05 & 54.57±1.18 & 64.27±0.37 \\
    Ours & 68.17±1.79 & 60.63±0.31 & 56.09±2.34 & 63.64±1.15 \\
    \bottomrule
  \end{tabular}
\end{table}

\begin{table}[htbp]
  \centering
  \caption{3-bit EWGS Results on OfficeHome Test Set (test)}
  \begin{tabular}{lcccc}
    \toprule
    Method & Art & Clipart & Product & Real-World \\
    \midrule
    EWGS & 26.98±0.51 & 36.88±3.95 & 48.00±3.38 & 44.48±3.86 \\
    EWGS + SAGM & 24.33±3.84 & 40.08±0.01 & 44.00±2.17 & 48.08±0.03 \\
    Ours & 31.57±2.52 & 41.60±0.84 & 47.02±3.27 & 47.78±1.19 \\
    \bottomrule
  \end{tabular}
\end{table}

\begin{table}[htbp]
  \centering
  \caption{3-bit LSQ Results on OfficeHome Validation Set (val)}
  \begin{tabular}{lcccc}
    \toprule
    Method & Art & Clipart & Product & Real-World \\
    \midrule
    LSQ & 66.84±3.48 & 53.22±0.11 & 56.37±0.78 & 61.90±3.03 \\
    LSQ + SAGM & 68.72±3.98 & 58.73±1.89 & 51.05±0.75 & 64.27±0.37 \\
    Ours & 67.36±2.53 & 61.88±0.49 & 53.71±1.30 & 65.53±0.13 \\
    \bottomrule
  \end{tabular}
\end{table}

\begin{table}[htbp]
  \centering
  \caption{3-bit LSQ Results on OfficeHome Test Set (test)}
  \begin{tabular}{lcccc}
    \toprule
    Method & Art & Clipart & Product & Real-World \\
    \midrule
    LSQ & 27.16±3.48 & 33.71±0.20 & 46.44±1.31 & 45.28±2.42 \\
    LSQ + SAGM & 30.74±4.74 & 40.48±1.39 & 40.61±0.72 & 48.08±0.03 \\
    Ours & 29.43±2.60 & 42.37±0.56 & 44.16±0.49 & 49.24±1.25 \\
    \bottomrule
  \end{tabular}
\end{table}